\documentclass[shortAfour,sageh,times]{sagej}

\usepackage[colorlinks,citecolor=black,urlcolor=black,filecolor=black,linkcolor=black,draft]{hyperref}
\usepackage[load=prefixed]{siunitx}
\usepackage{nth}
\usepackage{bm}
\usepackage{pbox}
\usepackage{algorithm}
\usepackage{tabularx}
\usepackage{multirow,graphicx}
\usepackage{hhline}
\usepackage{flushend} 
\usepackage{booktabs}
\usepackage{float}
\usepackage[inline]{enumitem}
\usepackage[fleqn]{amsmath}
\usepackage{subfig}
\usepackage{framed}
\usepackage{xcolor}

\DeclareMathOperator{\sgn}{sgn}
\DeclareMathOperator{\fmax}{fmax}

\setlist[enumerate,1]{%
  label=(\arabic*),
}

\makeatletter
 \def\@textbottom{\vskip \z@ \@plus 1pt}
 \let\@texttop\relax
\makeatother

\renewcommand{\eqref}[1]{Equation~(\ref{#1})}

\newcommand\BibTeX{{\rmfamily B\kern-.05em \textsc{i\kern-.025em b}\kern-.08em
T\kern-.1667em\lower.7ex\hbox{E}\kern-.125emX}}

\begin{document}

\onecolumn
{\Large

%
\noindent{Pre-print of article that will appear in the\\ \textbf{The International Journal of Robotics Research (IJRR)}.}\\

\noindent{Please cite this paper as:}\\
A. Valada, and W. Burgard, "Deep Spatiotemporal Models for Robust Proprioceptive Terrain Classification", \textit{The International Journal of Robotics Research (IJRR)}, 36(13-14):1521-1539, 2017.\\

\noindent{BibTex:}\\
\\
@inproceedings$\lbrace$valada17ijrr,\\
author = $\lbrace$Abhinav Valada and Wolfram Burgard$\rbrace$,\\
title = $\lbrace$Deep Spatiotemporal Models for Robust Proprioceptive Terrain Classification$\rbrace$,\\
journal = $\lbrace$The International Journal of Robotics Research (IJRR)$\rbrace$,\\
volume = $\lbrace$36$\rbrace$,\\
number = $\lbrace$13-14$\rbrace$,\\
pages = $\lbrace$15211--1539$\rbrace$,\\
doi = $\lbrace$10.1177/0278364917727062$\rbrace$,\\
year = $\lbrace$2017$\rbrace$ \\
publisher = $\lbrace$SAGE Publications Sage UK: London, England$\rbrace$,\\
$\rbrace$
}
\twocolumn

\runninghead{Valada and Burgard}

\title{Deep Spatiotemporal Models for Robust Proprioceptive Terrain Classification}
\author{Abhinav Valada and Wolfram Burgard}

\affiliation{Department of Computer Science, University of Freiburg, Germany \\ http://deepterrain.cs.uni-freiburg.de}

\corrauth{Abhinav Valada, Albert-Ludwigs-Universit\"at Freiburg,
Technische Fakult\"at,
Autonome Intelligente Systeme,
Georges-K\"ohler-Allee 079,
79110 Freiburg im Breisgau,
Germany.}

\email{valada@cs.uni-freiburg.de}

\begin{abstract}
\emph{Terrain classification is a critical component of any autonomous mobile robot system operating in unknown real-world environments. Over the years, several proprioceptive terrain classification techniques have been introduced to increase robustness or act as a fallback for traditional vision based approaches. However, they lack widespread adaptation due to various factors that include inadequate accuracy, robustness and slow run-times. In this paper, we use vehicle-terrain interaction sounds as a proprioceptive modality and propose a deep Long-Short Term Memory (LSTM) based recurrent model that captures both the spatial and temporal dynamics of such a problem, thereby overcoming these past limitations. Our model consists of a new Convolution Neural Network (CNN) architecture that learns deep spatial features, complemented with LSTM units that learn complex temporal dynamics. Experiments on two extensive datasets collected with different microphones on various indoor and outdoor terrains demonstrate state-of-the-art performance compared to existing techniques. We additionally evaluate the performance in adverse acoustic conditions with high ambient noise and propose a noise-aware training scheme that enables learning of more generalizable models that are essential for robust real-world deployments.}
\end{abstract}


\maketitle

\section{1. Introduction\label{sec:introduction}}

As the transition of robots from mere lab equipment to autonomous machines that tackle complex problems in unknown environments occurs, the perceptual challenges that they face increases exponentially. Mobile robots in particular should have the capability to distinguish the terrain that they traverse on and accordingly optimize their navigation strategy. Failing to adapt the trafficability can lead to disastrous situations. For instance, a robot that can travel with a certain maximum speed on carpet, cannot use the same speed to traverse on sand, if not it would lead to entrenchment. Moreover, the predominately adopted vision based approaches have a high probability of failure due to the similar visual appearance of these two terrains. Often, there are also situations where terrains are covered with leaves or water which can be very challenging for vision-based classifiers.

Over the years, this has motivated researchers to explore alternate modalities such as lidars \citep{thrun2006jfr, suger15icra}, vibrations induced on the vehicles body \citep{brooks2005ta}, vehicle-terrain interaction sounds \citep{libby2012icra, christie2016icra, cuneyitoglu2013mssp, valada2015isrr} and roughness estimation using accelerometers \citep{eriksson2008mobsys, weiss2006iros}. Each of these approaches have their own benefits and drawbacks: optical sensors perform remarkably well in the presence of good illumination but are drastically affected by changes in lighting conditions. Classification with RGB images is usually achieved using color and texture features extracted from the scene \citep{Sung2010jirs}. The use of texture-based image descriptors such as local ternary patterns have also been explored. In such approaches, features extracted using local ternary patterns are used on sequences of mutated images and classified using Recurrent Neural Network (RNN) configurations \citep{otte2015esann}. Lidars on the other hand, have extensively been used for traversability analysis. However, they are not suitable for fine-grained terrain classification where two or more terrains may have the same degree of coarseness. Nonetheless, they have been shown to benefit from being able to learn from partially labelled data and using semi-supervised learning techniques \citep{suger15icra, angel2015jfr}. Commonly, features used include statistics on remission values, roughness and slope. As there are benefits in using both images and lidar data, approaches have also been explored to learn classifiers from both modalities. Here, features from images include color, texture, and geometric features and from lidar data typically surface normals, curvature, ground height, point feature histograms, linearity and planarity are used \citep{namin2014iros,posner2008rss}. CNN-based approaches for both near-range and far-range terrain classification using stereo and RGB images have been demonstrated during the Defense Advanced Research Projects Agency (DARPA) Learning Applied to Ground Robots (LAGR) program \citep{hadsell2009jfr, muller2013spie}. These approaches achieved state of the art performance by combining both supervised and unsupervised learning using deep hierarchical networks. In contrast, acoustics based approaches perform well in all the above scenarios but are easily affected by environmental noise. Research using acoustics and other proprioceptive techniques are discussed in Section~2.

The benefit of exploring these alternate modalities is that the disturbances that affect optical or active sensors do not affect proprioceptive sensors such as microphones, hence enabling us to use them in a complementary classifier or fuse them with other modalities to increase robustness. These alternate approaches have failed to gain popularity beyond research. This can be attributed to the following:
\begin{enumerate*}
  \item manually designing feature sets that perform well in every real-world condition is tedious and impractical;
  \item existing techniques have slow run times making them unusable for real-time applications;
  \item most techniques require specific hardware setups that are difficult to replicate; and 
  \item approaches often lack reliability in real-world scenarios.
\end{enumerate*}

In this paper, we propose a novel multiclass terrain classification approach that overcomes these impediments by using only the sound from vehicle-terrain interaction to classify both indoor and outdoor terrains. We gathered two long-scale vehicle-terrain interaction datasets: one by equipping our mobile robot with a shotgun microphone and the other using a mobile phone microphone, as shown in Figure 1. The rationale behind this was to test the generalizability of the model to different hardware setups. Sounds from such interactions have distinct audio signatures and can be even used for fine-grained terrain classification such as distinguishing between grass and mowed-grass \citep{valada2015isrr}. Unlike speech or natural domestic sounds, vehicle-terrain interaction sounds are very unstructured in nature as several dynamic environmental factors contribute to the signal. Our previous work demonstrated that features learned using Deep Convolutional Neural Networks (DCNNs) significantly outperforms classification using shallow, manually-engineered feature sets. Our network combines three temporal pooling methods to achieve the time-series representation learning.

\begin{figure}
\begin{framed}
\centering
\includegraphics[scale=1]{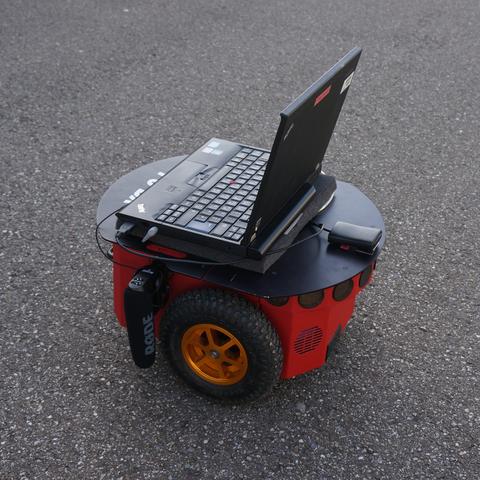}
\includegraphics[scale=1]{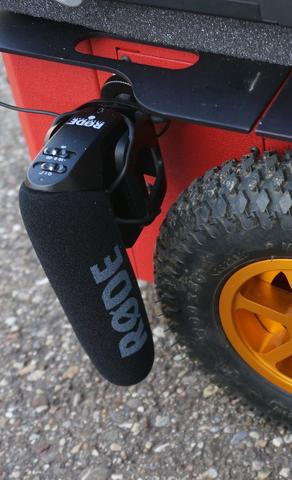}
\caption{The Pioneer P3-DX platform that we use in our experiments, showing the shotgun microphone with the shock mount mounted close to the wheel.}
\label{fig:platform}
\end{framed}
\end{figure}

In this work, we extend our previous approach by incorporating a deep LSTM recurrent framework in order to further exploit the temporal dynamics in our problem. Recurrent neural models form compositional representations in the time domain, similar to convolutional models forming compositional representations in the spatial domain. LSTM based recurrent neural network models have recently achieved impressive results for sequential learning tasks such as language translation \citep{sutskever2014nips}, speech recognition \citep{graves2014icml} and image captioning \citep{donahue2014cvpr}. The structure of our proposed network leverages spatio-temporal information in each clip and in transitioning clips as well, where we define a clip as the time window that we take for classification. We show that our end-to-end trained recurrent neural model learns improved feature representations compared to our DCNN on both our datasets. 

DCNNs learn highly discriminative features from training data, however, a caveat of this attribute is that the training data cannot encompass every possible real-world scenario. For acoustics-based terrain classification, this primarily relates to the ability of the model to adapt to ambient noise in the environment. Noise filtering techniques can be applied if the training data can include every possible noise disturbance, but this is highly impractical. Therefore, the model has to learn the distribution of common noises so that it can generalize effectively to different situations. Noise sensitivity is strongly correlated to the capability to operate in different real-world environments. Moreover, there are two different types of noise sources in our application; ambient noise and the noise from the motors of the robot. As the microphone is mounted close to the tires of the robot, it is inevitable that the captured signals also include the noise from the motors of the robot. Therefore this is already accounted for in the dataset. In order to make our model generalize effectively to real-world environments with varying types and amounts of ambient noise, we present a noise-aware training scheme. Our training scheme randomly injects ambient noise signals from the Diverse Environments Multichannel Acoustic Noise Database (DEMAND) \citep{demand2013} at different Signal-to-Noise (SNR) ratios. We then demonstrate that our recurrent model trained using our noise-aware scheme shows improved robustness in real-world conditions.

Specifically, the following are the main contributions of this paper:
\begin{enumerate}[nolistsep]
  \item We propose a new deep spatio-temporal architecture for learning complex dynamics in proprioceptive signals.
  \item We optimize various hyperparameters of the model and show their influence on the performance.
  \item We demonstrate that our approach is hardware independent and the performance of our model on data from an inexpensive hardware is on par with data from a high-quality device.
  \item We extensively evaluate the utility of our proposed Global Statistical Pooling (GSP) on various deep spatio-temporal model configurations.  
  \item We quantify our models performance in seven different environments having adverse acoustic conditions.
  \item We introduce a noise aware training scheme that substantially increases the generalizability of the model to real-world scenarios.
  \item We present thorough empirical evaluations on over $6\hour$ of audio data. 
\end{enumerate}

Rest of this paper is organized as follows. In Section~2, we discuss the work that has been previously done using acoustic and other proprioceptive sensors for terrain classification. We then detail our recurrent LSTM approach, the network architecture and training in Section~3. In Section~4, we provide a brief overview our data collection methodology. Finally we present our experimental evaluation in Section~5, followed by conclusions and discussion in Section~6.

\section{2. Related Work}
\label{sec:relatedworks}

Terrain classification using proprioceptive modalities has not been explored in the same depth as vision based approaches, yet there is a sizable amount of work in this area. The most researched proprioceptive terrain classification technique is using accelerometer data \citep{ojeda2006jfr, trautmann2011, weiss2006iros}. This is achieved by extracting features such as power spectral density, discrete fourier transform and other statistical measures from the vibrations induced on the vehicles body. Such approaches demonstrate a substantial amount of false positives for finer terrain classes such as asphalt and carpet. However, accuracies as high as $91.8\%$ has been reported for a seven class problem using Support Vector Machines (SVMs) \citep{weiss2006iros}. In a similar work, accelerometer data from a mobile sensor network system was used to detect potholes and other road anomalies \citep{eriksson2008mobsys}. Hand engineered features were used and the system achieved an average accuracy of 90\% in real-world experiments.

There is a body of work in terrain classification tailored to legged robots. Unlike in wheeled robots, proprioceptive terrain classification can enable safe foothold placement which is critical to ensure the stability of such legged systems. In one of the initial works \citep{hoepflinger2010}, the authors extracted features from ground contact forces and joint motor current measurements to train a multiclass AdaBoost classifier. Although they did not test the efficacy of their approach on real-world terrains, they demonstrated the classifiers capability to identify different coarseness and curvatures of surfaces, as a first step towards real-world proprioceptive terrain classification. In another approach \citep{best2013terrain}, the authors demonstrate the ability to classify four different outdoor terrains using position measurements from leg servos of a hexapod robot. They extract a 600-dimensional feature vector consisting of gait-phase domain, as well as frequency domain features. Their approach utilizes a $2.7\s$ window of data and a SVM is trained to classify the terrains.

Vibration data from contact microphones have also been successfully utilized for terrain classification. The vibrations captured are similar to accelerometers than air microphones that we use in this work, as they pick up only structure-borne sound and minimal environmental noise. Contact microphones have previously been mounted on analog rover's wheel frame to capture the induced vibrations and classify the terrain \citep{brooks2005ta}. They extract log-scale power spectral density features and train a pairwise classifier. For a three class problem, they achieve an average accuracy of $74\%$ on a wheel-terrain testbed and $85.3\%$ on a rover. They also demonstrate a self-supervised classification approach where a visual classifier is trained using labels provided by a vibration-based classifier \citep{brooks2007ac}.

The use of vehicle-terrain interaction sounds for terrain classification has been the most sparsely explored among all the proprioceptive modalities. Typically, combinations of state-of-the-art handcrafted audio features are used with traditional machine learning algorithms. Recently, a multiclass sound-based terrain classification system for a mobile robot was presented, that utilizes features including spectral coefficients, moments and various other spectral and temporal characteristics \citep{libby2012icra}. Their SVM based classifier achieves an average accuracy of $78\%$ over three terrain classes and three hazardous vehicle-terrain interaction classes. They also showed that smoothing over larger temporal window of about $2\second$ yields an improved accuracy of $92\%$. In another approach \citep{ojeda2006jfr}, a suite of sensors including microphones, accelerometers, motor current and voltage sensors, infrared, ultrasonics and encoders were used along with a feedforward neural network classifier. However their microphone based approach, only achieved an average accuracy of $60.3\%$ for a five class problem. They concluded that the overall performance was poor and such an approach was promising only for classes such as grass.

\begin{figure*}
\begin{framed}
\centering
\includegraphics[width=16cm,height=6.5cm]{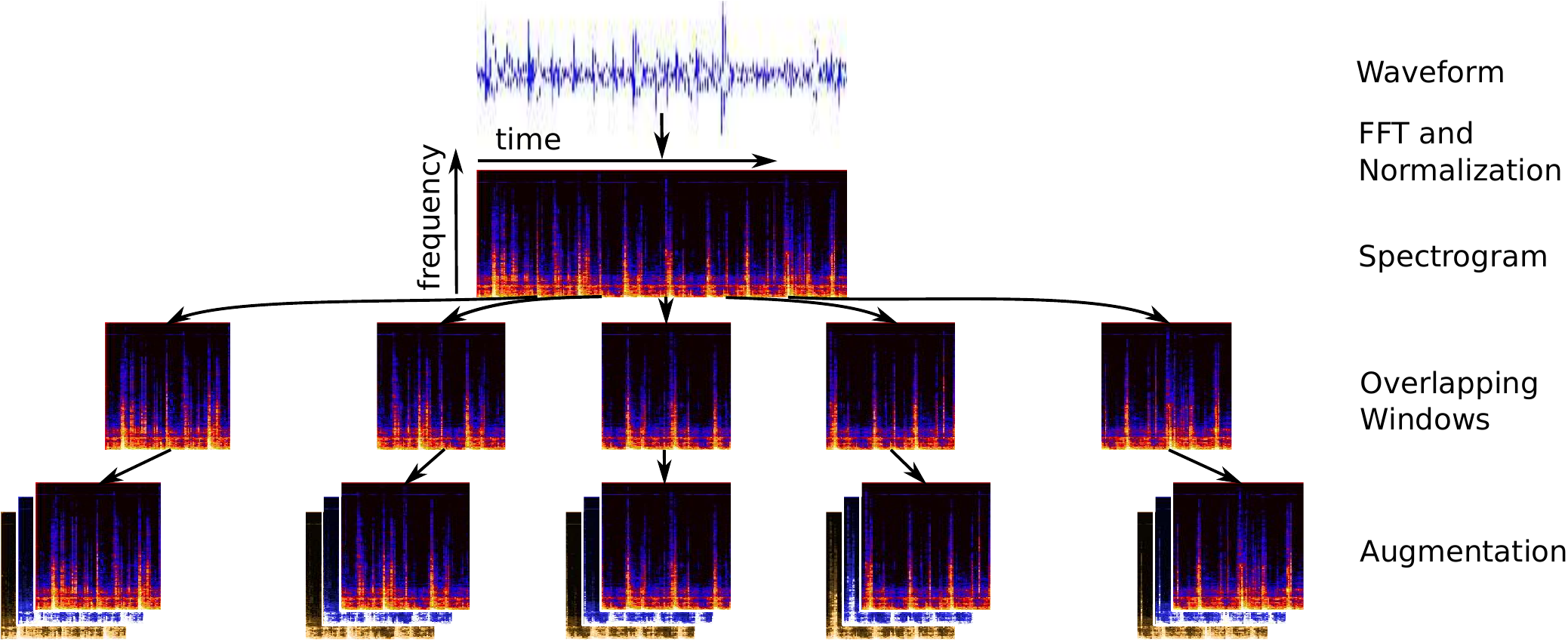}
\caption{Overview of our preprocessing pipeline. Raw audio waveform of the vehicle-terrain interaction is transformed into its spectrogram representation and then a series of augmentations are applied to forge the learned feature representation invariant to certain signal perturbations.}
\label{fig:preprocessing}
\end{framed}
\end{figure*}

Microphones have also been used for terrain classification with legged robots. In one such approach, terrain is classified from the steps taken by a hexapod robot \citep{cuneyitoglu2013mssp}. They use zero-crossing rate, frequency band coefficients and delta-features along with a functional trees classifier. They also experiment with different noise elimination techniques to remove motor/gear-head noise, but concluded that the performance was worse after the noise elimination. Their seven class classifier achieved an average accuracy of $90\%$. More recently, a SVM based system that uses statistics of spectral and band features, was used to classify terrain from the sounds produced during the locomotion of hexapod \citep{christie2016icra}. Their classifier was trained on a dataset that contains $5\minute$ worth of data from each terrain type. Their approach uses $1\second$ windows for classification and operates at $1\hertz$. They report an average accuracy of $89.08\%$ for seven terrain classes. They also investigated the use of spectral subtraction to eliminate the servo noise. They report an improvement in the sensitivity from $92.9\%$ to $95.1\%$ after spectral subtraction.

In all of the previous works, manually handcrafted features were extracted after specialized preprocessing steps. The approaches were evaluated on comparatively limited data and not in varying real-world environments. Most importantly, they do not model the temporally discriminative information in proprioceptive data which could potentially improve the classification performance, increase robustness and reduce false detections. In addition, they do not address the robustness of their models to ambient noise, which is one of the most critical properties for real-world adaptation. In the following section, we describe our DCNN architecture, the methodology using which we incorporate temporal recurrence into our model and our noise-aware training scheme.

\section{3. Technical Approach}
\label{technicalapproach}

One of the main objectives of our work is to develop an end-to-end trainable recurrent model tailored for classifying unstructured vehicle-terrain interaction sounds, with as minimal preprocessing as possible. Even though learning directly from raw waveforms has been demonstrated for speech processing, recent work has shown that the performance of such approaches is lower and the computational cost is higher, when compared to learning from simple transformations such as spectrograms \citep{khunarsal2013jis, graves2014icml}. Therefore we choose to train our models on top of this minimal preprocessing. This signal processing can also be computed in the input layer of the network. The features used for classification are learned by the network from this input representation. Figure 2 depicts our preprocessing pipeline.

\subsection{3.1. Spectrogram Transformation and Augmentation}

Unlike speech decoding or handwriting recognition tasks that require specific alignment between the audio and transcription sequences, our application does not necessitate a specific target for each frame. We only need to classify a set of frames/clips without requiring any prior alignment in the sequences. Therefore, we use individual clips as a new sample for classification. We first split the raw audio signal into short clips of $t_w$ and $t_{ov}$ of overlap between the clips. We then experimentally identify the shortest clip length, overlap and the number of windows taken by the LSTM units that best enables our model to learn the discriminative information from the time varying vehicle-terrain interaction signal.

We extract the Short-Time Fourier Transform (STFT) based spectrogram for each clip in the sequence. We first block each audio clip into $M$ samples with $50\%$ overlap between the frames. Let $x[n]$ be the recorded raw audio signal with duration of $N_f$ samples, $f_s$ be the sampling frequency, $S(i,j)$ be the spectrogram representation of the 1-D audio signal and $f(k)=k f_s/ N_f$. By applying STFT on the length $M$ windowed frame of the signal, we get
\begin{equation}
\begin{split}
X(i,j) = \sum\limits_{p=0}^{N_{f}-1} x[n]\;w[n-j]\exp{\left(-p\frac{2 \pi k}{N_f}n\right)},\\
p = 0,\ldots,N_{f}-1
\end{split}
\end{equation}
We use a Hamming window function $w[n]$ to compensate for the Gibbs effect while computing STFT by smoothing the discontinuities at the beginning and end of the audio signal, i.e.,
\begin{equation}
\begin{split}
w[n] = 0.54-0.46 \cos\left(2\pi \frac{n}{M-1}\right),\\
n = 0,\ldots,M-1
\end{split}
\end{equation}
We then compute the log of the power spectrum as 
\begin{align}
S_{\log}(i,j) = 20 \log_{10}(|X(i,j)|)
\end{align}

We chose $N_f$ as $2048$ samples, therefore the spectrogram contains $1024$ Fourier coefficients. Experiments from our previous work \citep{valada2015isrr} revealed that most of the spectral energy is concentrated below $512$ coefficients, therefore we only use the lower $512$ coefficients to reduce the computational complexity and runtime. The noise and intensity levels vary a fair amount in the entire dataset as we collected data in several real-world environments. Factors such as environmental noise at different times of the day, variations in weather conditions and variations in the servo noise at different speeds, contribute to this. Therefore, we normalize the spectrograms by dividing by the maximum amplitude. We then compute the mean spectrum over the entire dataset and subtract it from the normalized spectrogram. We compute this as
\begin{align}
S(i,j) = S_{\log}(i,j)/\max_{i,j} S_{\log}(i,j)
\end{align}

\begin{figure*}
\begin{framed}
\centering
\includegraphics[width=16cm,height=6cm]{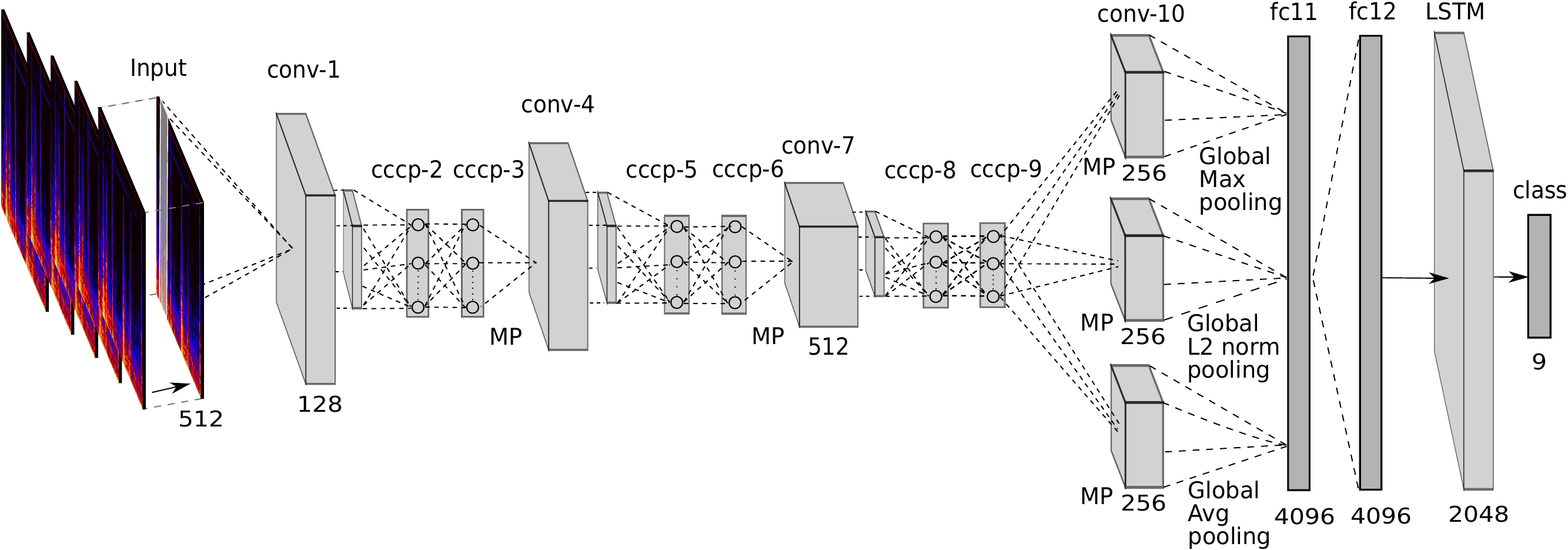}
\caption{Overview of our terrain classification pipeline. We first transform the raw audio signal of the terrain interaction into its spectrogram representation and then feed it into a DCNN for feature learning and classification. The LSTM is unrolled in time and MP refers to max pooling.}
\label{fig:network}
\end{framed}
\end{figure*}

Data augmentation has been shown to improve the performance of several deep learning tasks and also help in training of very deep networks. We experimented with several signal augmentation strategies and created additional samples by applying a set of augmentations $A_t$ on the audio signal in the frequency domain. Using two dimensional affine transform and warping, random offsets in time and frequency were applied to perform shifting of the spectrogram. Furthermore, we created more samples using time stretching, modulating the tempo, random equalization, and by varying the volume gain. We also experimented with frequency and time normalization with a sliding window and local contrast normalization.
   
\subsection{3.2. Network Architecture}

Recurrent Neural Network (RNN) models learn complex temporal dynamics from a sequence by mapping inputs $x = (x_1,\ldots,X_T)$, to a sequence of hidden states $h=(h_1,\ldots,h_T)$, and hidden states to an output sequence $y=(y_1,\ldots,y_T)$. This is achieved by iterating through the following equations from $t=1 \text{ to } T$, as

\begin{align}
h_t &= g(W_{ih}x_t + W_{hh}h_{t-1} + b_h) \\
y_t &= W_{ho}h_t + b_o
\end{align}

where $g$ is is the hidden layer activation function such as an element-wise application of sigmoid non-linearity, $W$ and $b$ terms denote the weight matrices and bias vectors, with subscripts $i, h, \text{ and } o$ denoting input, hidden and output respectively.

As discussed extensively in previous works \citep{hochreiter1997lstm, graves2014icml, donahue2014cvpr}, traditional RNNs often suffer from vanishing and exploding gradient problems that occur from propagating gradients through several layers of the RNN. The larger the length of temporal input, the harder it is to train the RNN. The long short-term memory (LSTM) architecture \citep{hochreiter1997lstm} was proposed as a solution to this problem and to enable exploitation of long-term temporal dynamics from a sequence. LSTMs incorporate memory units containing several gates that regulate the flow of information in and out of the cells.

We first use our DCNN architecture \citep{valada2015isrr}, to extract features from the spectrogram of each clip in the sequence in time. The spectrograms in our training set are of the form $S = \{s_1,\ldots,s_T\}$ with $s_{i} \in \mathbb{R}^{N}$. Each of them are of size $v \times w$ and number of channels $d$ ($d=1$ in our case). Our network, shown in Figure 3, consists of six convolution layers, six Cascaded Cross Channel Parametric Pooling (CCCP) layers \citep{lin2013arxiv}, three Fully-Connected (FC) layers, an LSTM layer and $\mathit{softmax}$ layer. All the convolution layers are one dimensional with a kernel size of three and convolve along the temporal dimension. We use a fixed convolutional stride of one. CCCP layers that follow the first, second and third convolution layers are used to enhance discriminability for local patches within receptive fields. CCCP layers effectively employ $1{\times}1$ convolutions over the feature maps and the filters learned are a better non-linear function approximator. A max-pooling layer with a kernel of two follows the second, fourth and sixth CCCP layers. Max-pooling adds some invariance by only taking the high activations from adjacent hidden units that share the same weight, thereby providing invariance to small phase shifts in the signal.

DCNNs that are designed to operate on images for various perception tasks, specifically preserve the spatial information of features learned. However, our application does not benefit from this, as we are only interested in identifying the presence or absence of features, rather than localizing the features in the frame. To this end, we introduce a new Global Statistical Pooling (GSP) scheme that applies different pooling mechanisms across entire input feature maps and combines them to gather statistics of features across a particular dimension. Our architecture utilizes this GSP by incorporating three different global pooling layers after cccp-9 to compute the statistics of the features across time. We use an inner product layer to combine the outputs of max pooling, L2 norm pooling and average pooling. In our previous work \citep{valada2015isrr}, we investigated various combinations of different global pooling layers and found that for other configurations, the accuracy dropped over $3\%$. As our new proposed architecture utilizes LSTM units to model the temporal relationships, we explore the utility of this global statistics pooling in Section~5. We use the Xavier weight filler \citep{glorot2010aistats} to initialize the convolution and CCCP layers by drawing from a zero mean uniform distribution from $[-a,a]$ and the variance as a function of input neurons $n_{in}$, where $a = \sqrt{3\mathbin{/}n_{in}}$. The effects of initializing with just a Gaussian filler are discussed in Section~5.

Rectified linear units (ReLUs) significantly help in overcoming the vanishing gradient problem. We use ReLUs $f(x) = \max(0,x)$, after the convolution layers and dropout regularization \citep{hinton2012arxiv} on the inner product layers. We then stack the LSTM layer after fc12, followed by another inner product layer and a $\mathit{softmax}$ layer. If $x_t$ is the input to the LSTM layer at time $t$, the activations can be formulated as
\begin{align}
i_t &= \sigma (W_{xi} x_t + W_{hi} h_{t-1} + b_i) \\
f_t &= \sigma (W_{xf} x_t + W_{hf} h_{t-1} + b_f) \\
o_t &= \sigma (W_{xo} x_t + W_{ho} h_{t-1} + b_o) \\
g_t &= \phi (W_{xc} x_t + W_{hc} h_{t-1} + b_c) \\
c_t &= f_t \odot c_{t-1} + i_t \odot g_t \\
h_t &= o_t \odot \phi(c_t)
\end{align}
where $\sigma(x) = (1 + e^{-x})^{-1}$ is the sigmoid nonlinearity and $\phi(x) = \frac{e^{x} - e^{-x}}{e^{x} + e^{-x}} = 2\sigma(2x)$ is the hyperbolic tangent nonlinearity. $h_t \in \mathbb{R}^{N}$ is the hidden unit, $g_t \in \mathbb{R}^{N}$ is the input modulation gate, $c_t \in \mathbb{R}^{N}$ is the memory cell, $i_t \in \mathbb{R}^{N}$ is the input gate, $f_t \in \mathbb{R}^{N}$ is the forget gate, and $o_t \in \mathbb{R}^{N}$ is the output gate. $W$ and $b$ are the weight matrix and bias with subscripts $i, f, h, c, \text{and } o$ representing input, forget, hidden, cell and output gates respectively. $\odot$ represents element-wise multiplication. The hidden state $h_t$ models the terrain that the robot is traversing on at time $t$. The output of the memory cell changes over time based on the past states and the current state of the cell. Therefore, the hidden state is formed based on the short-term memory of the past clip. At timestep $t$, the predicted distribution $P(y_t)$ can be computed by taking the $\mathit{softmax}$ over the outputs of the sequence from the LSTM units $z_t$, i.e.,

\begin{figure}
\begin{framed}
\centering
\includegraphics[scale=0.5]{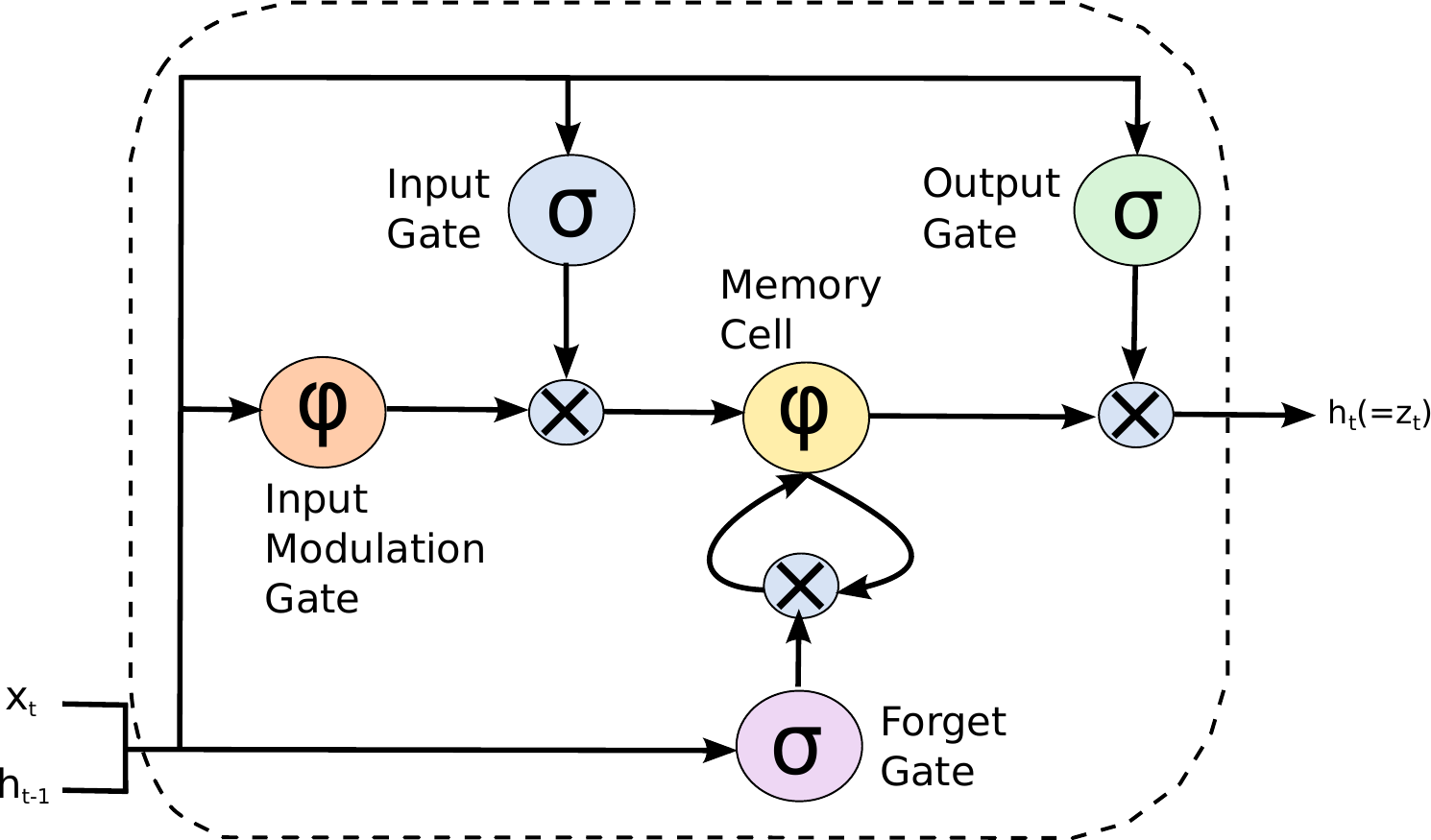}
\caption{Depiction of a Long Short-term Memory Cell \citep{zaremba2014ariv, graves13arxiv}.}
\label{fig:lstm}
\end{framed}
\end{figure}

\begin{align}
P(y_t=c) = \frac{\text{exp}(W_{zc} z_{t,c} + b_c)}{\sum\limits_{c' \in C} \text{exp}(W_{zc} z_{t,c'} + b_c)}
\end{align}

We train the entire model end-to-end using minibatch Stochastic Gradient Decent (SGD) with a batch size of $256$. We optimize the SGD by smoothing the gradient computation for minibatch using a momentum coefficient $\alpha$ as $0 < \alpha < 1$. The update rule can then be written as

\begin{align}
\Delta w_{ij}(t) = \alpha \Delta w_{ij}(t-1)-\epsilon \frac{\partial E}{\partial w_{ij}(t)}
\end{align}

We found the model to be extremely sensitive to the learning parameters used. Therefore, we use the Spearmint Bayesian optimization library \citep{snoek2012nips} to tune the hyperparameters. We first optimized the learning policy and the initial learning rate over fixed, inverse, step and poly policies. We obtained the best performance using an initial learning rate of $\lambda_0 = 0.01$ and the poly policy as $\lambda_n = \lambda_0 \times (1-N/N_{\max})^c$, where $N$ is the iteration number, $N_{\max}$ is the maximum number of iterations, and $c$ is power. We tuned the number of outputs in the inner product layers and the LSTM layer as they were highly correlated and had a big impact on the performance of the model. We increased the number of parameters in the inner product layers in order to make the LSTM model converge. This parameter tuning is further discussed in Section~5.

\subsection{3.3. Noise Aware Training}

The basic principle of our noise aware training scheme is to randomly inject common environmental noises into the training data so that the signal features are learned along with noise patterns, and hence accounted for during inference. As our hierarchical model encompasses a spatial and a temporal component, the lower spatial layers learn features that describe both the pure signal and noise signal, whereas the higher layers learn to distinguish between them. The temporal component enables the model to learn the evolution of both the signals. Due to the spatio-temporal depth, the network is capable of learning heterogeneous signal patterns and at what stage to de-emphasize the noise while making decisions. Noise-conditioned decision boundaries are leaned as we train the network with signals corrupted with noise at different Signal-to-Noise Ratios (SNRs). The noise-aware training scheme also regularizes the network. It helps improve the classification of pure signals as the easily degraded parts of the signal are blurred by noise which forces the network to learn the more dominant features, hence avoiding over-fitting. 

During the training stage, we randomly select a noise following a multinomial distribution $\mathrm{Mult}(\mu_1, \mu_2,\ldots,\mu_n)$, where $n$ is the types of noises and $\mu_i$ is sampled from a Dirichlet distribution as $(\mu_1, \mu_2,\ldots,\mu_n)~\sim~\mathrm{Dir}(\alpha_1,\alpha_2,\ldots,\alpha_n)$, where $\alpha_i$ is set to control the base distribution of the noise types. Furthermore, the SNR of the noised sample follows a Gaussian distribution $\mathcal{N}(\mu_{\mathrm{SNR}}, \sigma_{\mathrm{SNR}})$, where $\mu_{\mathrm{SNR}}$ and $\sigma_{\mathrm{SNR}}$ are the mean and variance. We randomly select a start point $s$ on the noise signal and scale it according to a SNR before adding it to the pure vehicle-terrain clip. The noise samples are concatenated if the noise clip length is smaller than the vehicle-terrain clip length. We use the Praat framework \citep{praat2013} for the noise corruption. 

We use $18$ classes of noise recordings from the Diverse Environments Multichannel Acoustic Noise Database (DEMAND). We categorize the noise into seven classes for in-depth experiments as follows:

\begin{itemize}[noitemsep]
\item White: White noise has a very wide band and it is one of the most common noise sources. It has a very similar effect to that of various physical and environmental disturbances including wind and water sources. 
\item Domestic: This category includes noises from living rooms, kitchens and washing rooms. As our terrain classes also contain indoor terrains, it is critical to train the model with common indoor ambient noises.
\item Nature: The nature category contains outdoor noise samples from a sports field, a river creek with flowing water and a city park.
\item Office: This category contains recordings from an office with people working on computers, a hallway where people pass by occasionally.and from interior public spaces.
\item Public: The public category contains recordings from interior public spaces such a bus station, a cafeteria with people and at a university restaurant.
\item Street: The street category contains noise recordings from outdoor inner-city public roads and pedestrian traffic. It contains a busy traffic intersection, a town square with tourists and a cafe at a public space.
\item Transportation: This category contains recordings of vehicle noises such as cars, subways and buses.
\end{itemize}

As vehicle-terrain sounds are already unstructured in nature, corrupting it with various heterogeneous noises might lead to the model not converging. We first investigate the effect of these disturbances on models trained for audio-based terrain classification. We then train our network with a fixed SNR on all the above ambient noise categories and compare it with training on corrupted samples with varying SNRs. The weights and biases for the noise-aware training are initialized by copying them from our pure trained model as described in the previous section. We use a learning rate $1/10th$ of the initial rate used for training the network. Results from these experiments are discussed in section~5.4.

\subsection{3.4. Baseline Feature Extraction}

There is a wide spectrum of audio features developed for various audio recognition and detection tasks. In previous works, researchers have explored the utilization of these features for sound-based terrain classification \citep{libby2012icra, cuneyitoglu2013mssp, christie2016icra}. In order to evaluate the performance of our proposed model, we extracted several of these traditional baseline audio features. We explored the use of both time and frequency domain features. For wheeled mobile robots, previously, Ginna and Shape features have been demonstrated to yield the best results \citep{libby2012icra}.  Ginna features \citep{giannakopoulos2006setn} is a six dimensional feature vector consisting of zero crossing rate (ZCR), short time energy (STE), spectral centroid (SC), spectral rolloff (SR) and spectral flux (SF). While, shape features \citep{wellman1997spie} is four dimensional feature vector consisting of spectral centroid, standard deviations, skewness and kurtosis.

\begin{figure*}
\begin{framed}
\centering
\setlength{\tabcolsep}{0.2em}
\begin{tabular}{p{2.4cm} p{2.4cm} p{0.2cm} p{2.4cm} p{2.4cm} p{0.2cm} p{2.4cm} p{2.4cm}}
\includegraphics[height=2.4cm,width=2.4cm]{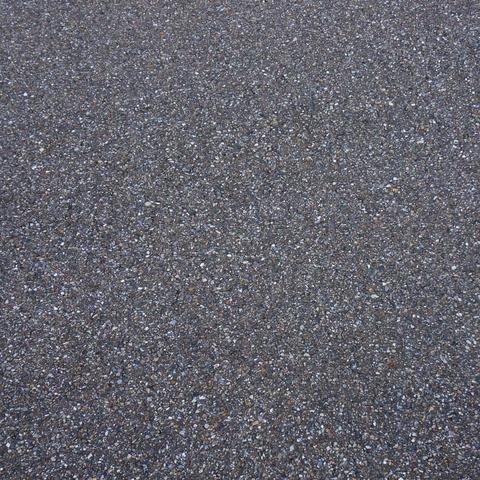} & 
\includegraphics[height=2.4cm,width=2.4cm]{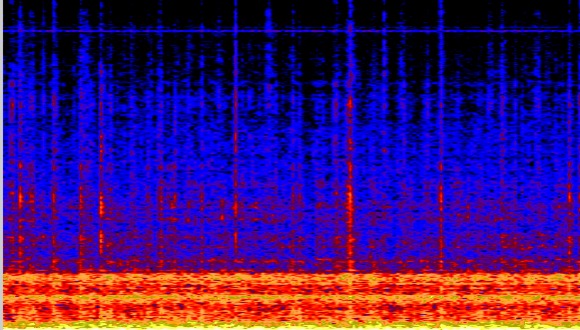} & & \includegraphics[height=2.4cm,width=2.4cm]{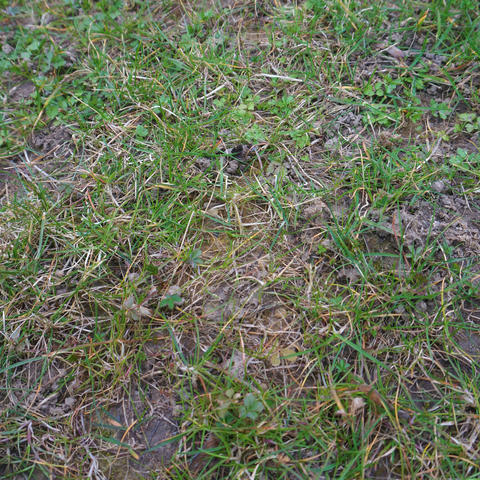} & \includegraphics[height=2.4cm,width=2.4cm]{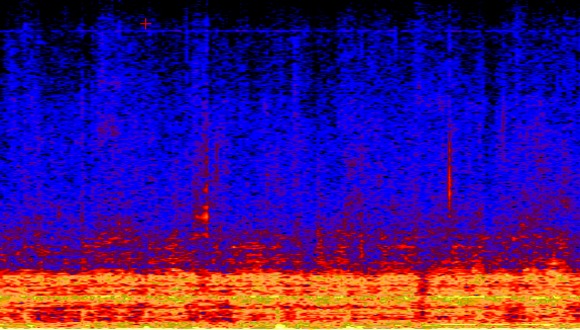} & & \includegraphics[height=2.4cm,width=2.4cm]{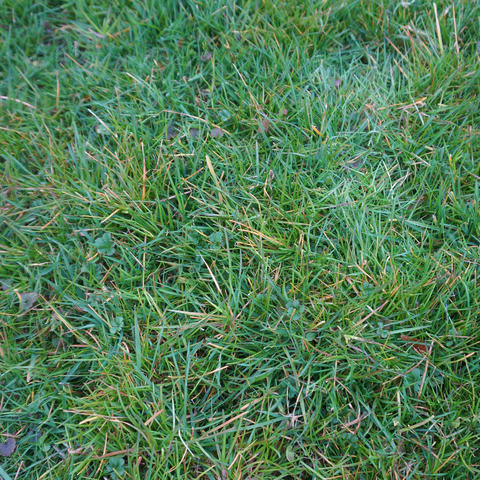} & 
\includegraphics[height=2.4cm,width=2.4cm]{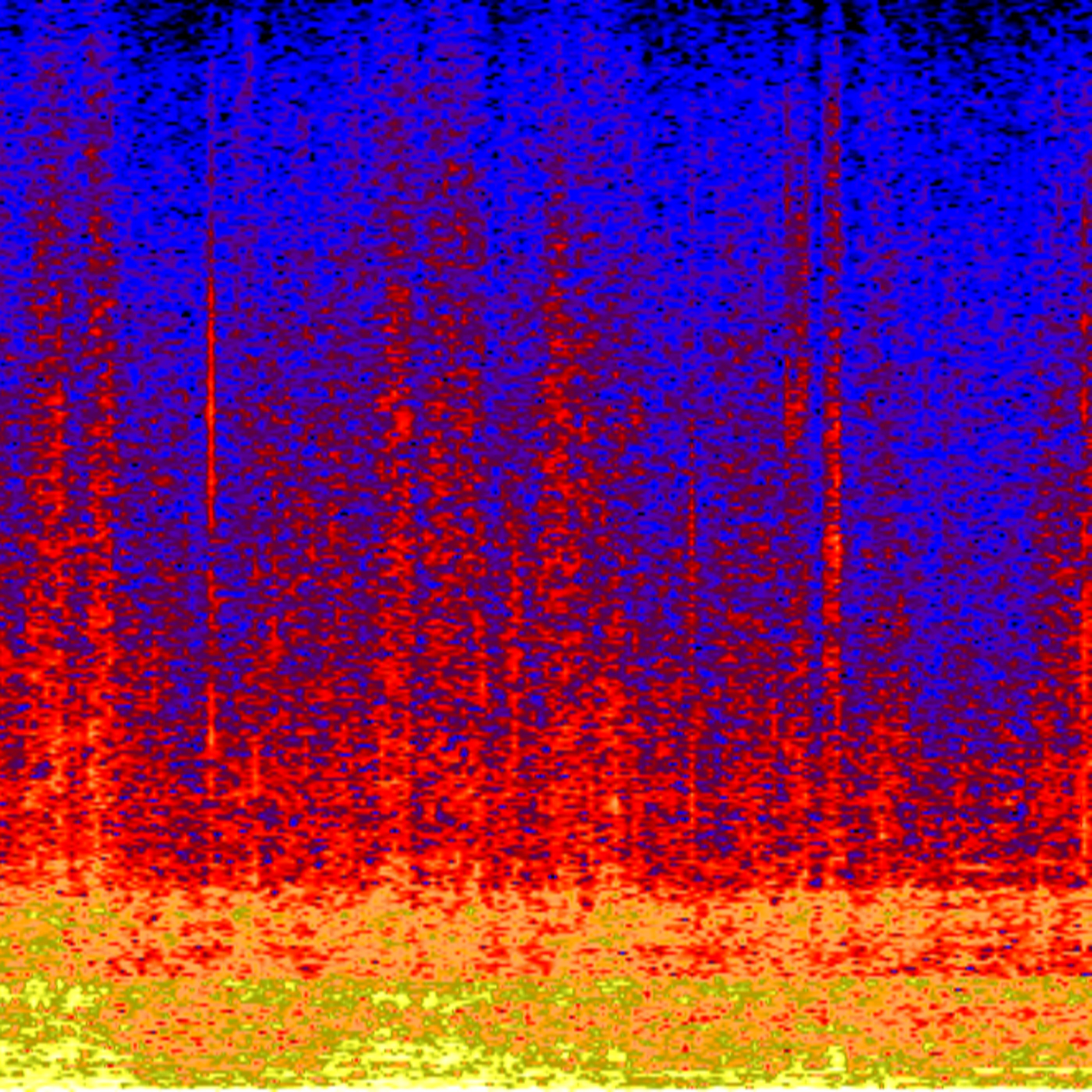} \\
\multicolumn{2}{l}{(a) Asphalt} & & \multicolumn{2}{l}{(b) Mowed Grass} & & \multicolumn{2}{l}{(c) Grass Med-High} \\[6pt]
\includegraphics[height=2.4cm,width=2.4cm]{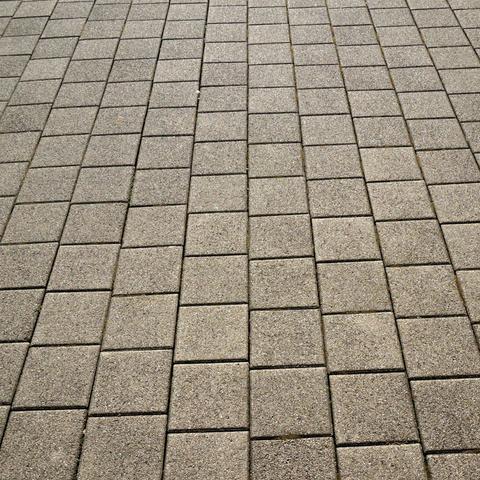} & 
\includegraphics[height=2.4cm,width=2.4cm]{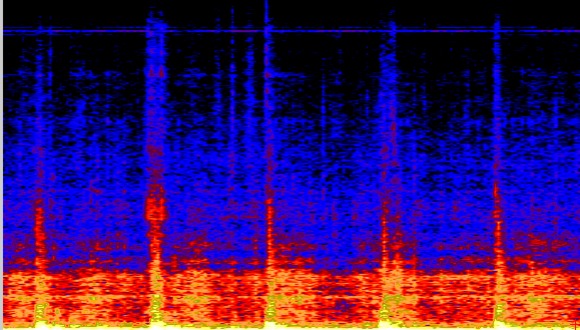} & &
\includegraphics[height=2.4cm,width=2.4cm]{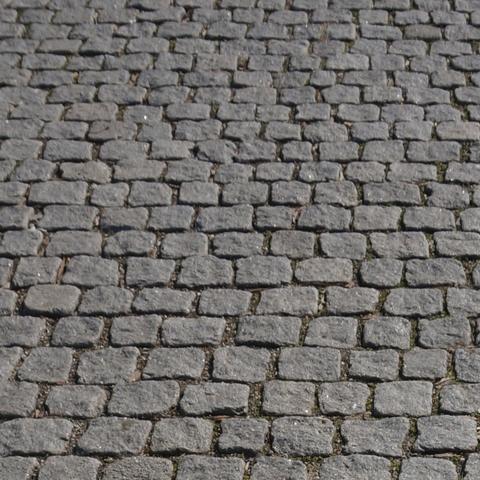} & 
\includegraphics[height=2.4cm,width=2.4cm]{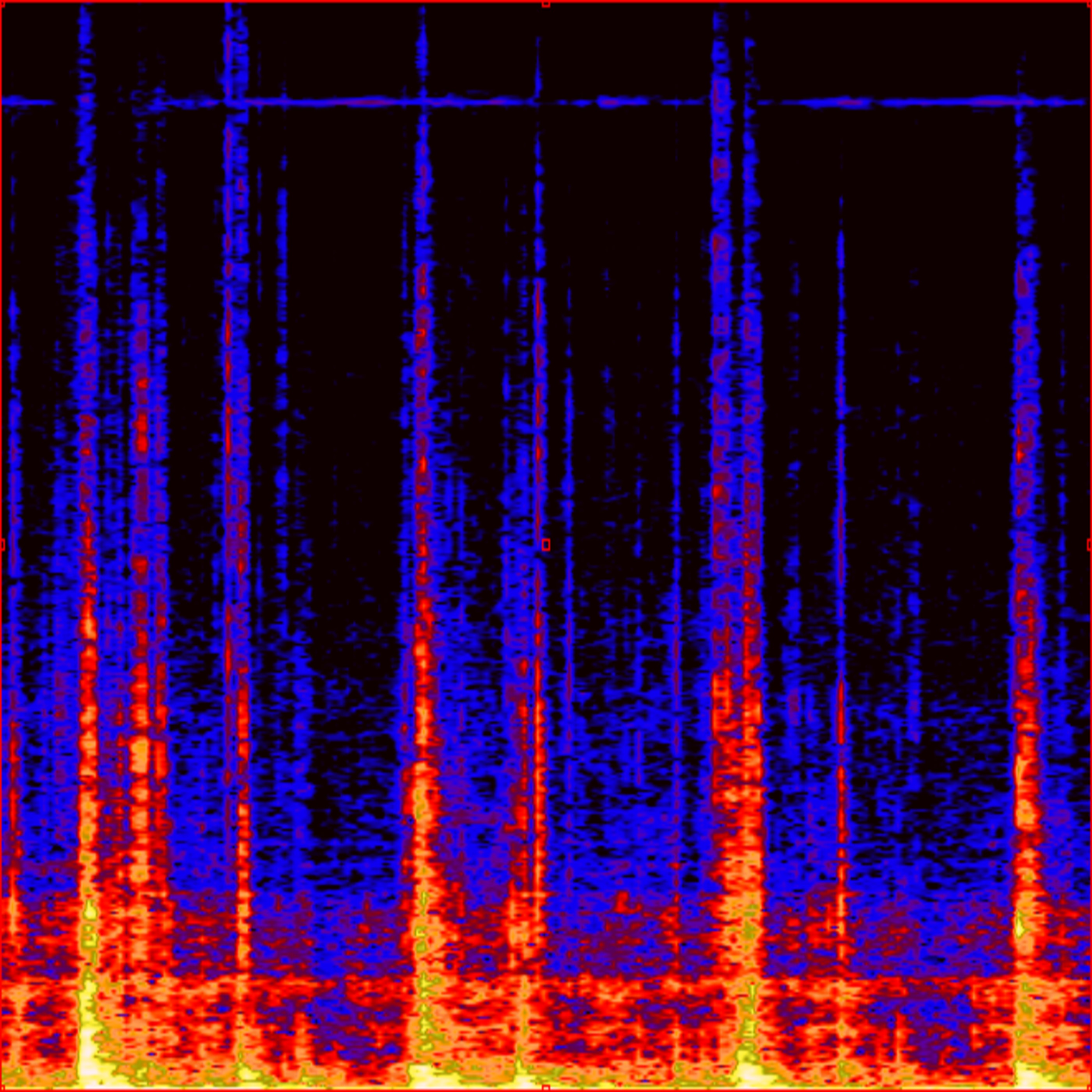} & &
\includegraphics[height=2.4cm,width=2.4cm]{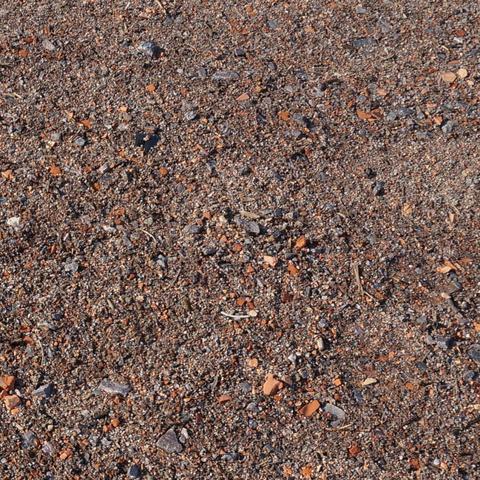} & 
\includegraphics[height=2.4cm,width=2.4cm]{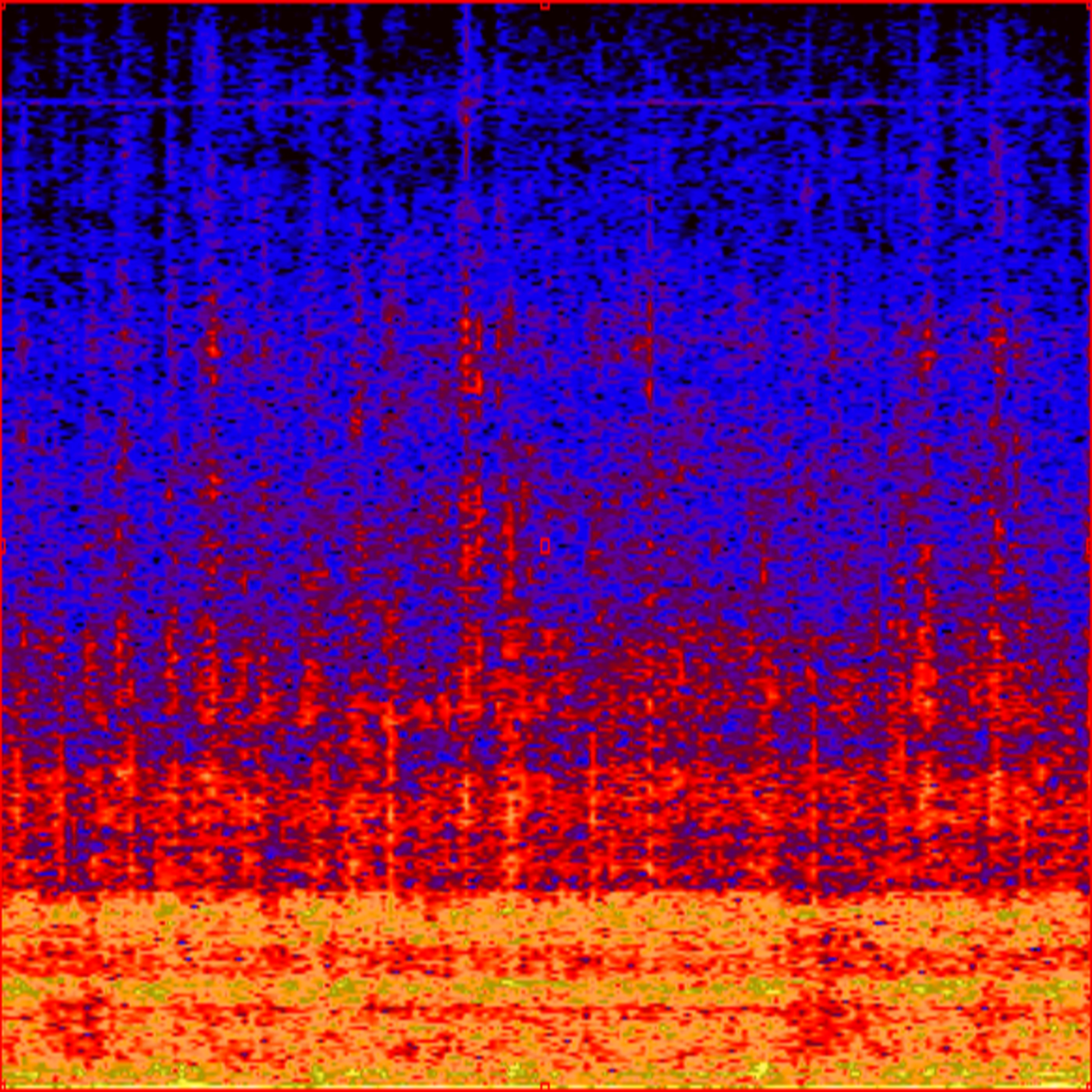} \\
\multicolumn{2}{l}{(d) Paving} & & \multicolumn{2}{l}{(e) Cobble} & & \multicolumn{2}{l}{(f) Offroad} \\[6pt]
\includegraphics[height=2.4cm,width=2.4cm]{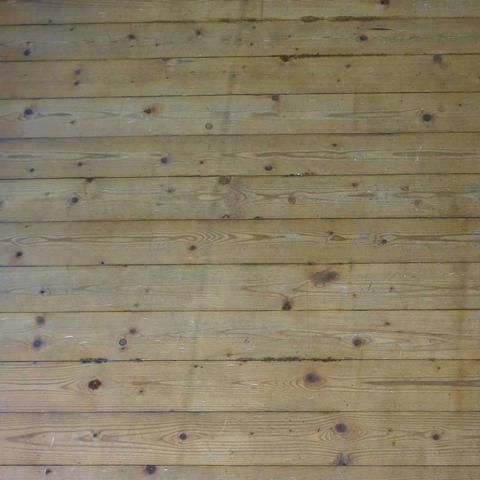} & 
\includegraphics[height=2.4cm,width=2.4cm]{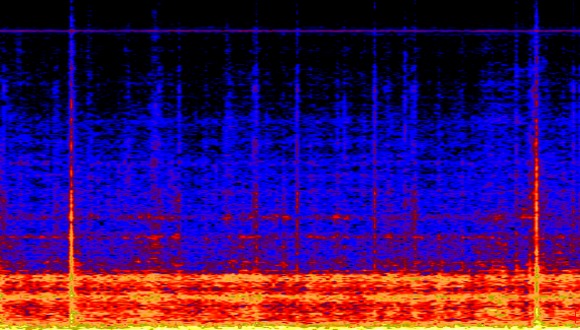} & &
\includegraphics[height=2.4cm,width=2.4cm]{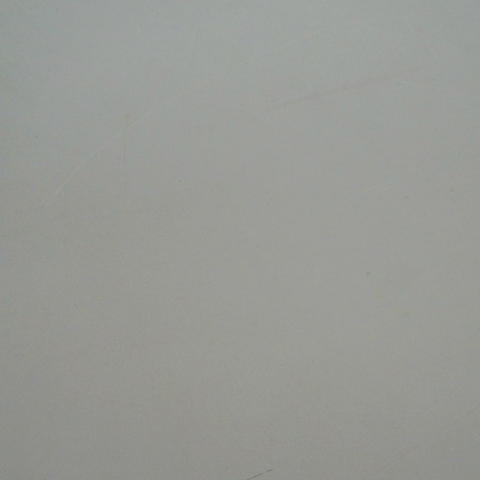} & 
\includegraphics[height=2.4cm,width=2.4cm]{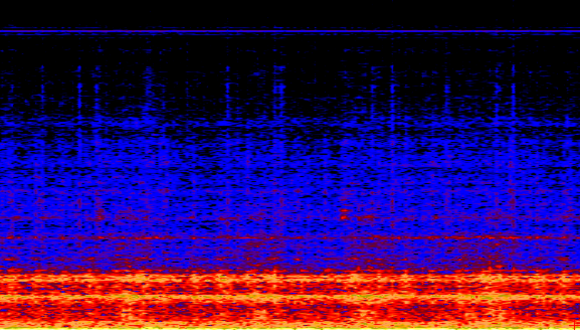} & &
\includegraphics[height=2.4cm,width=2.4cm]{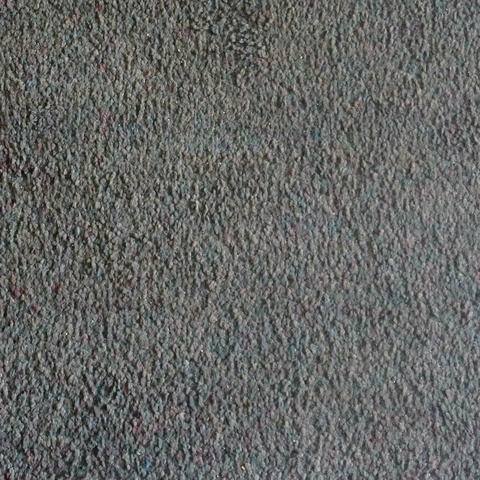} & 
\includegraphics[height=2.4cm,width=2.4cm]{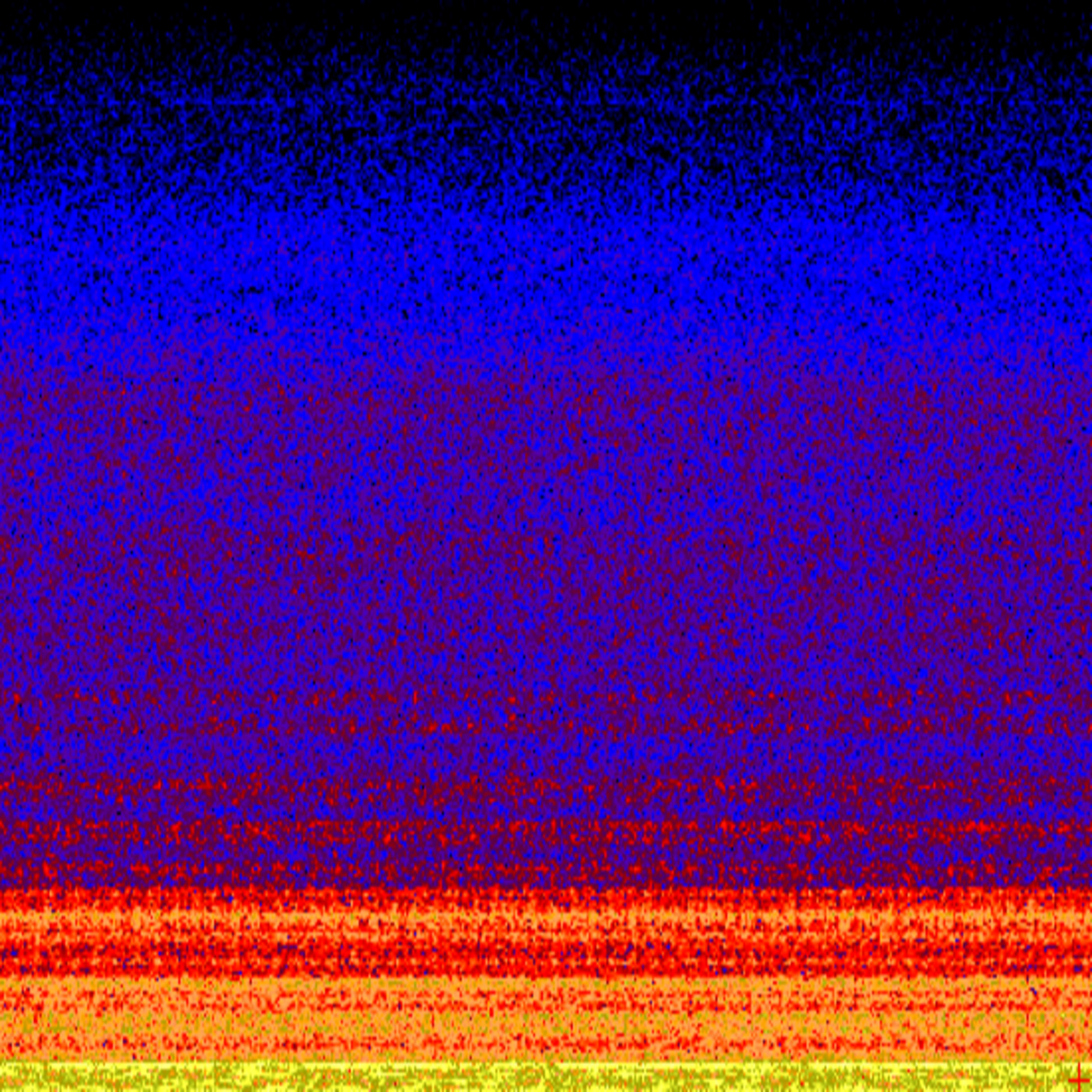} \\
\multicolumn{2}{l}{(g) Wood} & & \multicolumn{2}{l}{(h) Linoleum} & & \multicolumn{2}{l}{(i) Carpet} \\[6pt]
\end{tabular}
\caption{Terrain classes and an example of their corresponding spectrogram for a $2,000\milli\second$ clip (colorized spectrograms are only shown for better visualization, spectrograms used for training are one channel).}
\end{framed}
\end{figure*}

ZCR can be defined as the rate at which the signal changes from positive to negative. ZCR is one of the most traditional audio feature used for speech and music classification. STE can be defined as the energy in short segment of the signal. It can be given as the square of the amplitude of the signal. Spectral centroid also referred to as the median of the power spectrum, indicates where the center of mass of the spectrum is. They can be calculated as
\begin{align}
\mathrm{ZCR} &= \frac{1}{2N} \sum\limits_{m=0}^{N-1} |\sgn[y(m)] - \sgn[y(i-1)]| \\
\text{where, } &\sgn[x(m)] = \begin{cases}
1, x(n) \geq 0 \\
-1, x(n) < 0
\end{cases} \\
\mathrm{STE} &= \sum\limits_{m=0}^{N-1} X((m).w(n-m))^2 \\
\mathrm{SC} &= \frac{\sum\limits_{m=0}^{N-1} f(m) X(m)}{ \sum\limits_{m=0}^{N-1} X(m)}
\end{align}
where, $X(n)$ is a sample of the power spectrum at bin $n$, $f(n)$ is the center frequency of the bin $n$, $w(n)$ is the windowing function, and $N$ is the size of the window. Spectral flux indicates the change in the power spectrum between successive windows. Spectral rolloff is the frequency below the $95\text{th}$ percentile of the power in the spectrum. They can be given as
\begin{align}
\mathrm{SF} &= \frac{\sqrt{\sum\limits_{m=1}^{N-1}\left(X(m,n)-X(m,n-1)\right)^2}}{N-1} \\
\mathrm{SR} &= K, \text{where} \sum\limits_{m=0}^{K} X(m) = 0.95 \sum\limits_{m=0}^{\fmax} X(m)
\end{align}
where $\fmax$ is the maximum frequency at bin $m$. Spectral skewness measures the symmetry of the distribution of the spectral magnitude around the mean. Spectral kurtosis measures the similarity of the distribution of the spectral magnitude to that of a Gaussian distribution. They can be defined by
\begin{align}
\mathrm{SS} &= \frac{1}{S} \sum\limits_{m=0}^{N-1} \left(\frac{X(m)- \mu}{\sigma}\right)^3 \\ 
\mathrm{SK} &= \frac{1}{S} \sum\limits_{m=0}^{N-1} \left(\frac{X(m)-\mu}{\sigma}\right)^4 - 3 \\
\text{where } \mu &= \frac{1}{S} \sum\limits_{m=0}^{N-1} X(m) \\
\sigma &= \sqrt{\frac{1}{S} \sum\limits_{m=0}^{N-1} (X(m)-\mu)^2} 
\end{align}

where, $\mathrm{S}$ is half the size of the window $S = N/2$, $\mu$ is the mean across the windows, and $\sigma$ is the standard deviation across the windows. A combination of Mel-frequency Cepstral Coefficients (MFCCs) and Chroma features have been demonstrated to yield state of the art performance for music classification tasks \citep{ellis2007}. MFCCs are one of the most widely used features for audio classification and Chroma features are strongly related to the harmonic progression of the audio signal. We use a combination of 12-bin MFCCs and 12-bin Chroma features for our comparisons. Timbral features containing means and variances of spectral centroid, spectral rolloff, spectral flux, zero crossing rate, low energy, and first 5 MFCCs, have shown impressive results for music genre classification \citep{tzanetakis2002tsap}. We also use this 19 dimensional feature vector for our baseline comparison. For our final baseline comparison, we use  a combination of cepstral features containing 13-bin MFCCs, Line Spectral Pair (LSP) and Linear Prediction Cepstral Coefficients (LPCCs) \citep{verma2008pattern}.

\section{4. Data Collection and Labelling}
\label{datacollection}

\begin{table*}
\centering
\caption{Classification accuracy of several baseline feature extraction approaches on our dataset for a clip length of $300\milli\second$. Our recurrent neural model performs $8.83\%$ better than the best baseline approach.}
\label{tab:featuresAndClassifiers}
\begin{tabular}{p{5cm}p{3.5cm}p{3.5cm}p{3.5cm}}
\toprule
Features & SVM Linear & SVM RBF & k-NN \\
\noalign{\smallskip}\hline\hline\noalign{\smallskip}
Ginna & 44.87 $\pm$ 0.70 & 37.51 $\pm$ 0.74 & 57.26 $\pm$ 0.60 \\
Spectral & 84.48 $\pm$ 0.36 & 78.65 $\pm$ 0.45 & 76.02 $\pm$ 0.43 \\
Ginna \& Shape & 85.50 $\pm$ 0.34 & 80.37 $\pm$ 0.55 & 78.17 $\pm$ 0.37 \\
MFCC \& Chroma & 88.95 $\pm$ 0.21 & \textbf{88.55} $\bm{\pm}$ \textbf{0.20} & 88.43 $\pm$ 0.15 \\
Trimbral & 89.07 $\pm$ 0.12  & 86.74 $\pm$ 0.25 & 84.82 $\pm$ 0.54 \\
Cepstral & \textbf{89.93} $\bm{\pm}$ \textbf{0.21} & 78.93 $\pm$ 0.62 & \textbf{88.63} $\bm{\pm}$ \textbf{0.06} \\
\midrule
DCNN (ours) & & {\textbf{97.52} $\bm{\pm}$ \textbf{0.016}} & \\
DCNN with LSTM (ours) & & {\textbf{98.76} $\bm{\pm}$ \textbf{0.009}} & \\
\bottomrule
\end{tabular}
\end{table*}

To the best of our knowledge, audio-based terrain classification using mobile robots has been focused on outdoor terrains. Whereas, there has been some work on both indoor and outdoor audio-based terrain classification using legged robots. Our objective is to create a model that can classify a wide variety of both indoor and outdoor terrains. Some indoor and outdoor terrains can have very similar visual features (Figure~5(a), 5(h), 5(i)) and hence pose a challenge for classifying using the vision based counterparts. We collected over $15 \hour$ of vehicle-terrain interaction data from six different outdoor terrains and three different indoor terrains.

Our approach is platform independent, as the network learns features from the given training data. For our experiments, we use the P3-DX mobile robot platform to collect data as it has a small footprint and feeble motor noise. We equipped the P3-DX with rugged wheels that enabled us to collect data both indoors and structured as well as unstructured outdoors environments. One of the big challenges for audio classification is to deal with ambient and environmental noise. Training the network with data that includes interference from nearby sound sources can drastically affect the performance. In order to prevent such biases in training data, we equip the robot with a shotgun microphone to record vehicle-terrain interaction sounds. A shotgun microphone has a supercardioid polar pickup pattern, which helps in rejecting off-axis ambient sounds. Specifically, we chose the Rhode VideoMic Pro and mounted it near the wheel of the mobile robot as shown in Figure 1. The microphone has an integrated shock mount that prevents the pickup of any unwanted vibrations caused during the traversal.

We collected data in several different locations to have enough variability and to enable our model to generalize effectively to real-world environments. Therefore even signals in the same class have varying spectral and temporal characteristics. The robots speed was varied from $0.1 \si{\meter\per\second}$ to $1.0 \si{\meter\per\second}$ during the data collection runs. The data was recorded in the lossless 16-bit WAV format at $44.1 \si{\kilo\hertz}$ to avoid any recording artifacts. There was no software level boost added during the recordings as we found it to amplify the ambient noise substantially, instead we used a $20 \si{\decibel}$ hardware level boost. The data was manually labelled using the live tags and timestamps made during the recordings. A waveform analyzer tool was used to crop out significant noise disturbances that had substantially higher amplitudes than the vehicle-terrain interaction signals. By noisy disturbances, we refer to uncommon temporary environmental disturbances such as a nearby car or train passing. The data is then split to train and test sets, ensuring that the classes have approximately the same number of samples to prevent any bias towards a specific class. It was also ensured that the training and validation sets do not contain clips from the same location.

\section{5. Experimental Results}
\label{experimentalresults}

We use the Caffe \citep{jia2014arxiv} deep learning framework for our implementations and the LSTM described in \citep{zaremba2014ariv} but a version of the implementation faster than \citep{donahue2014cvpr}. All our models are trained end-to-end and the experiments were run on a system containing a NVIDIA TITAN X GPU with cuDNN acceleration. The results from our experiments are described in the following sections.

\subsection{5.1. Baseline Comparisons}

In this section, we empirically evaluate our base DCNN model and our LSTM model with several baseline audio features discussed in Section~4.3. For all the baseline experiments that we present in Table~1, we choose a fixed clip length of $300\si{\milli\second}$. We compare with SVMs and k-Nearest Neighbours (kNNs) classifiers, which have shown the best performance in the work by \citep{libby2012icra}. SVMs perform well in high dimensional spaces and kNNs perform well when there are irregular decision boundaries. As a preprocessing step for the baseline features, we normalize the data to have zero mean. We use the {one-vs-all} voting scheme with SVMs to handle multiple classes and experimented with Linear and Radial Basis Function (RBF) as decision functions. We use inverse distance weighting for kNNs and optimized the hyperparameters for both classifiers by a grid-search using cross-validation. For the baseline classifiers, we used the implementations in scikit-learn and LibSVM. The results from this comparison are showed in Table~1.

The best performing baseline feature combination was Cepstral features using a linear SVM kernel, achieving an accuracy of $89.93\%$. The performance of Trimbral features using a linear SVM kernel is comparable. Ginna and Shape features using an SVM RBF kernel was the best performing feature set in the work of \citep{libby2012icra}. It achieved an accuracy of $80.37\%$, which is $9.56\%$ less than our best performing baseline. The worst performance was using only Ginna features with an SVM RBF kernel. Table~1 shows that the features containing MFCCs outperform other feature combinations. The feature sets used also contain the mean and standard deviations of each feature.

Our DCNN yields an accuracy of $97.52\%$, which is an improvement of $7.59\%$ over the best performing Cepstral features and $12.02\%$ over Ginna and Shape features. This is currently the state of the art for acoustics-based terrain classification for a clip length of $300 \si{\milli\second}$. Our recurrent LSTM model achieves an improved accuracy of $98.76\%$ with the same clip length and an LSTM window of $3$. This demonstrates that learning spatio-temporal relationships can further improve the classification performance. Note that this is not the best result of our network; model analysis and parameter tuning, which further improve the model, are discussed in the following sections. Our DCNN model performs inference in $9.15\si{\milli\second}$ and our recurrent LSTM model performs inference in $12.37\si{\milli\second}$ for a clip length of $300 \si{\milli\second}$, whereas baseline approaches presented in Table~1 have feature extraction and classification time in the order of a few seconds.

\begin{figure}
\begin{framed}
\centering
\includegraphics[width=\linewidth]{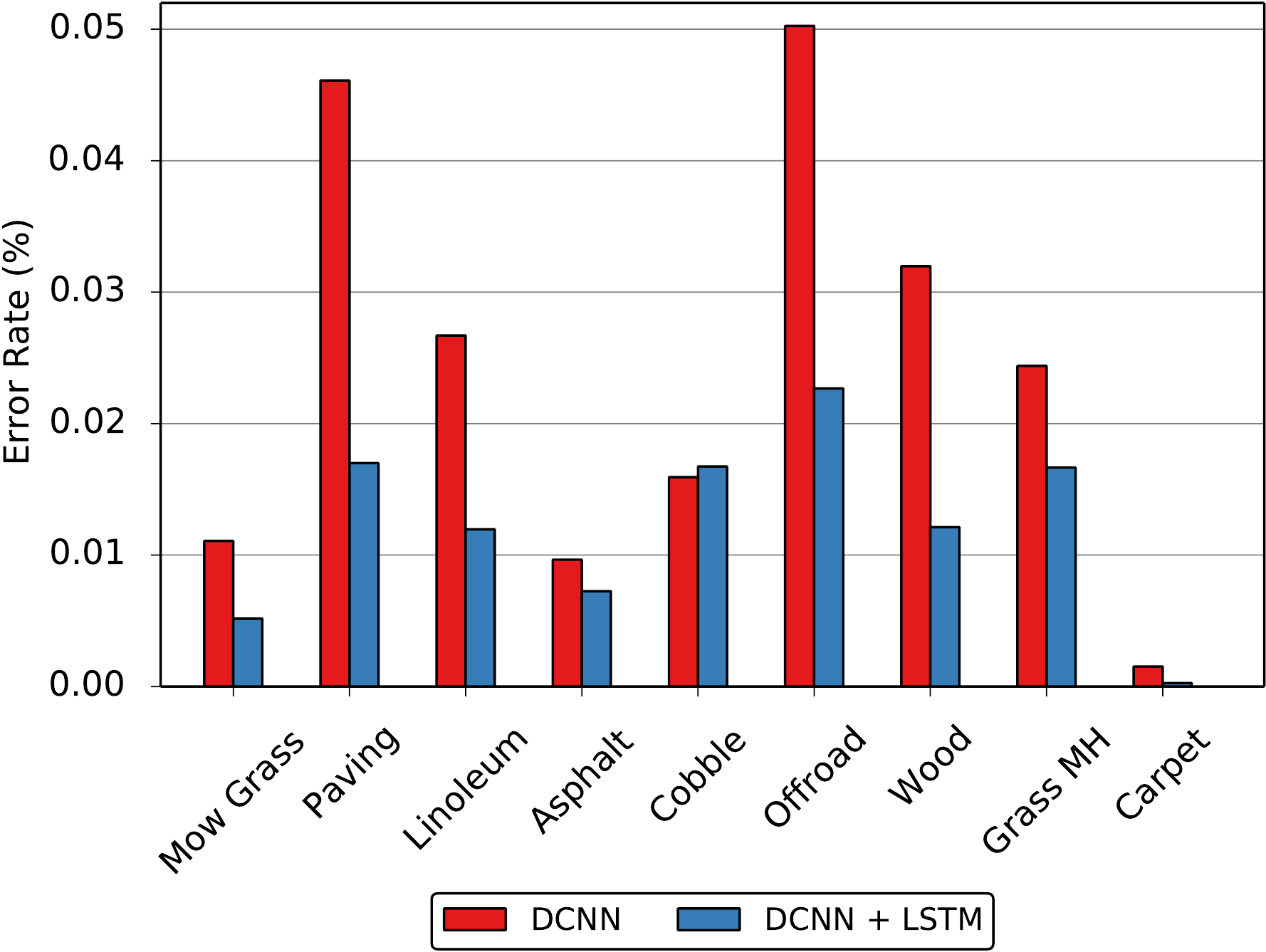}
\caption{Per-class error rate of our DCNN and our new proposed DCNN with LSTM model, for a clip length of $300\si{\milli\second}$. Our recurrent neural model achieves $1.24\%$ decrease in error rate compared to our DCNN model.}
\label{fig:errorcompDcnnLstm}
\end{framed}
\end{figure}
   
Moreover, Figure~6 shows the comparison of error rates of our DCNN model and DCNN with LSTM model. For all the classes other than Cobble, there is a considerable decrease in the error rate. For the Cobble class, there is an insignificant increase of $0.0008\%$ in error. Paving and Offroad classes have the largest decrease in error. In our previous work, we showed that these two classes have the highest false positives and the results now demonstrate that even this complex relationship can be learned using our temporally recurrent model.

\subsection{5.2. Network Parameter Estimation}

In this section, we describe our experiments that we perform to gain insight on the effect of learning spatio-temporal relationships using different DCNN model configurations. We first describe each of the models and then present an in-depth analysis of their performance. For consistency, we use the same $300\si{\milli\second}$ clip length, similar to the baseline experiments.

We consider the following four spatio-temporal model configurations:

\begin{enumerate}
\item M1 (DCNN): This model has a structure similar to standard classification models where there are alternating convolution and pooling layers which help in learning spatial features across the frames of a clip, followed by an inner product and softmax layer. The convolutions are only one dimensional and convolve along the temporal dimension.  
\item M2 (DCNN with GSP): This model is similar to the M1 model described above but it incorporates our global statistical pooling before the inner product layer, which helps in learning statistics of temporal features across time.
\item M3 (DCNN with LSTM): This model is a temporal extension of the M1 model described above. An LSTM layer is appended after the first two inner product layers to learn complex temporal dynamics across clips. The LSTM layer is followed by one inner product layer and a softmax layer.
\item M4 (DCNN with GSP \& LSTM): This model is a temporal extension of the M2 model described above. Global statistical pooling is first used to combine statistics of pooled features across time, then the resulting features are fed to an LSTM module to learn temporal dynamics across several clips. Similar to M3, the LSTM layer is followed by one inner product layer and a softmax layer. We call this model DCNN~ST (Spatio-temporal) for future comparisons.
\end{enumerate}

\begin{table}
\centering
\caption{Comparison of our deep spatio-temporal model configurations. Our temporally recurrent model demonstrates $7.47\%$ improvement compared to its spatial DCNN counterpart.}
\label{tab:modelbase}
\begin{tabular}{p{4.3cm}p{0.7cm}p{0.7cm}p{0.7cm}}
\toprule
Model & Acc. & Prec. & Rec. \\
\noalign{\smallskip}\hline\hline\noalign{\smallskip}
M1 (DCNN) & 91.29 & 91.88 & 91.56 \\
M2 (DCNN + GSP) & 97.52 & 97.56 & 97.61 \\
M3 (DCNN + LSTM) & 95.73 & 95.93 & 95.88 \\
M4 (DCNN + GSP \& LSTM) & \textbf{98.76} & \textbf{98.75} & \textbf{98.82} \\
\bottomrule
\end{tabular}
\end{table}

We use the same train and test splits for the $300\si{\milli\second}$ clips as in Valada~\textit{et al.}, 2015. For the LSTM models, we use a window size of three. Results from this comparison are shown in Table~2. Interestingly, the DCNN with GSP model outperforms the DCNN with LSTM model by $1.79\%$. This can be attributed to the fact that the LSTM model does not benefit much from learning using large clip lengths and short temporal windows. In the later part of this section, we investigate the influence of clip lengths and temporal window lengths. However, the DCNN with GSP and LSTM outperforms all the other models by achieving an accuracy of $98.76\%$, which is a $7.47\%$ improvement over the performance of the DCNN model. This also illustrates that GSP is critical for learning effective spatio-temporal relations in vehicle-terrain interaction sounds.

Figure~7 shows the per-class sensitivity of the spatio-temporal models discussed. The performance of the DCNN with GSP and LSTM model surpassed all the other models for every class other than Grass Medium-High. The performance of the DCNN with GSP model is better than that of the DCNN with GSP and LSTM model for the Grass Medium-High class. All the models have near perfect sensitivity for the Carpet class. This is primarily because the spectral responses for this class is mostly flat while compared to the others. The LSTM models perform better for flatter terrains such as Wood, Carpet and Paving, while the GSP models perform better when the terrains are more irregular. The DCNN model without the temporal features perform substantially worse than the others for almost all the classes.

\begin{figure}
\begin{framed}
\centering
\includegraphics[width=\linewidth]{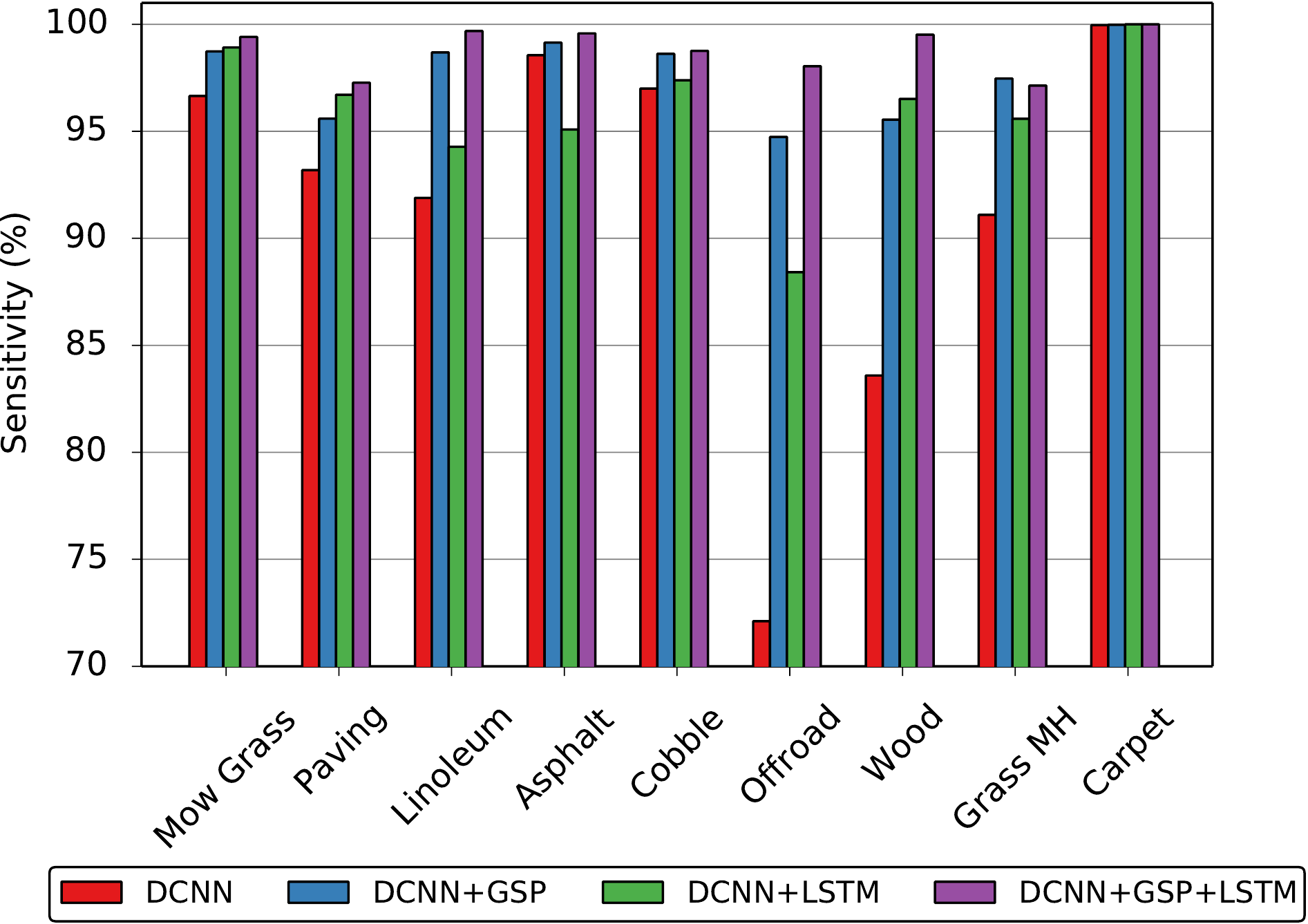}
\caption{Comparison of the per-class sensitivity of the spatio-temporal model configurations. Our temporally recurrent model outperforms the other configurations in almost all the classes.}
\label{fig:recallSpatioTemp}
\end{framed}
\end{figure}

\begin{table}
\centering
\caption{Classification accuracy of our DCNN~ST model for varying LSTM window lengths and audio clip lengths. Accuracy in \%.}
\label{tab:lstmwindow}
\begin{tabular}{p{1cm}ccccc}
\toprule
Window Size & 2 & 3 & 4 & 5 & 6 \\
\noalign{\smallskip}\hline\hline\noalign{\smallskip}
$300\si{\milli\second}$ & 97.43 & \textbf{98.76} & 80.09 & 98.70 & 98.62 \\
$250\si{\milli\second}$ & 96.74 & 97.38 & \textbf{98.93} & 79.96 & 98.67 \\
$200\si{\milli\second}$ & 96.69 & 96.22 & 97.86 & \textbf{99.03} & 79.58 \\
\bottomrule
\end{tabular}
\end{table}

Moreover, there are a number of critical parameters that are required to be optimized for efficient spatio-temporal learning. The biggest trade-off is with selecting the optimal clip length. As each clip is a new sample for classification, the shorter the clip length, the higher is the rate at which we can infer the terrain. A larger clip length yields increased accuracy, although this leads to an increase in the execution time and decrease in the classification rate, which are both undesirable. For our application, fast classification and execution rates are essential for making quick trafficability decisions. In our previous work \citep{valada2015isrr}, we showed that a clip length of $300\si{\milli\second}$ was sufficient considering the above trade-offs. However, the trade-off now has another degree of complexity, as the LSTM window size also needs to be optimized. Nonetheless, using LSTMs to learn temporal relationships allows us to use a shorter clip length. We investigate the relationship between these parameters by individually training models with various clip lengths and LSTM window lengths. Table~3 shows the results from these experiments.

\begin{table}
\centering
\caption{Performance comparison of our DCNN model (M2) and DCNN~ST model (M4) at varying audio clip lengths and the corresponding time taken to process though the pipeline. Accuracy in \% and time in $\milli\second$.}
\label{tab:windowvsfeature}
\begin{tabular}{ccccc}
\toprule
Clip Length & \multicolumn{2}{c}{DCNN} & \multicolumn{2}{c}{DCNN~ST} \\
\cmidrule(lr){2-3} \cmidrule(lr){4-5}
 & {Accuracy} & {Time} & {Accuracy} & {Time} \\
\midrule
\midrule
2000 ms & 99.86 & 45.40 & 99.88 & 23.27 \\ 
1500 ms & 99.82 & 34.10 & 99.83 & 21.16 \\
1000 ms & 99.76 & 21.40 & 99.78 & 16.43 \\
500 ms & 99.41 & 13.30 & 99.45 & 13.75 \\
300 ms & 97.36 & 9.15 & 98.76 & 12.37 \\
250 ms & 94.05 & 9.15 & 98.93 & 12.36 \\
200 ms & 91.30 & 9.14 & 99.03 & 12.23 \\
\bottomrule
\end{tabular}
\end{table}

\begin{figure}
\begin{framed}
\centering
\includegraphics[width=\linewidth]{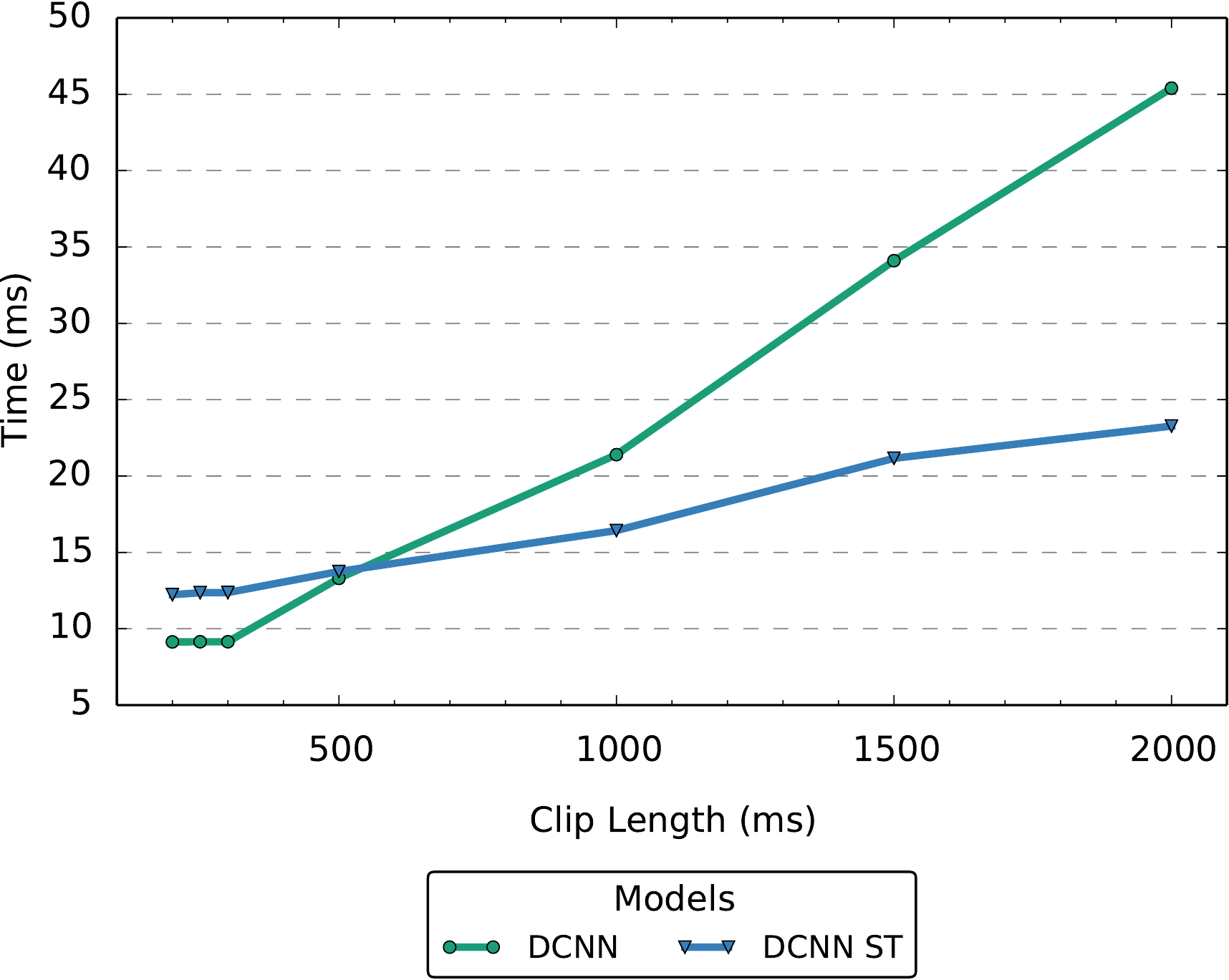}
\caption{Comparison of classification rates of the DCNN model and DCNN~ST model at varying audio clip lengths. Our DCNN~ST model has comparatively faster classification rates for a specific clip length.}
\label{fig:dcnnVsDcnnst}
\end{framed}
\end{figure}

\begin{figure*}
\begin{framed}
\centering
\subfloat[Confusion matrix of our DCNN model (M2) for a clip length of $300\si{\milli\second}$ \citep{valada2015isrr}\label{subfig-1:cm_300ms}\vfill]{%
\includegraphics[scale=0.26]{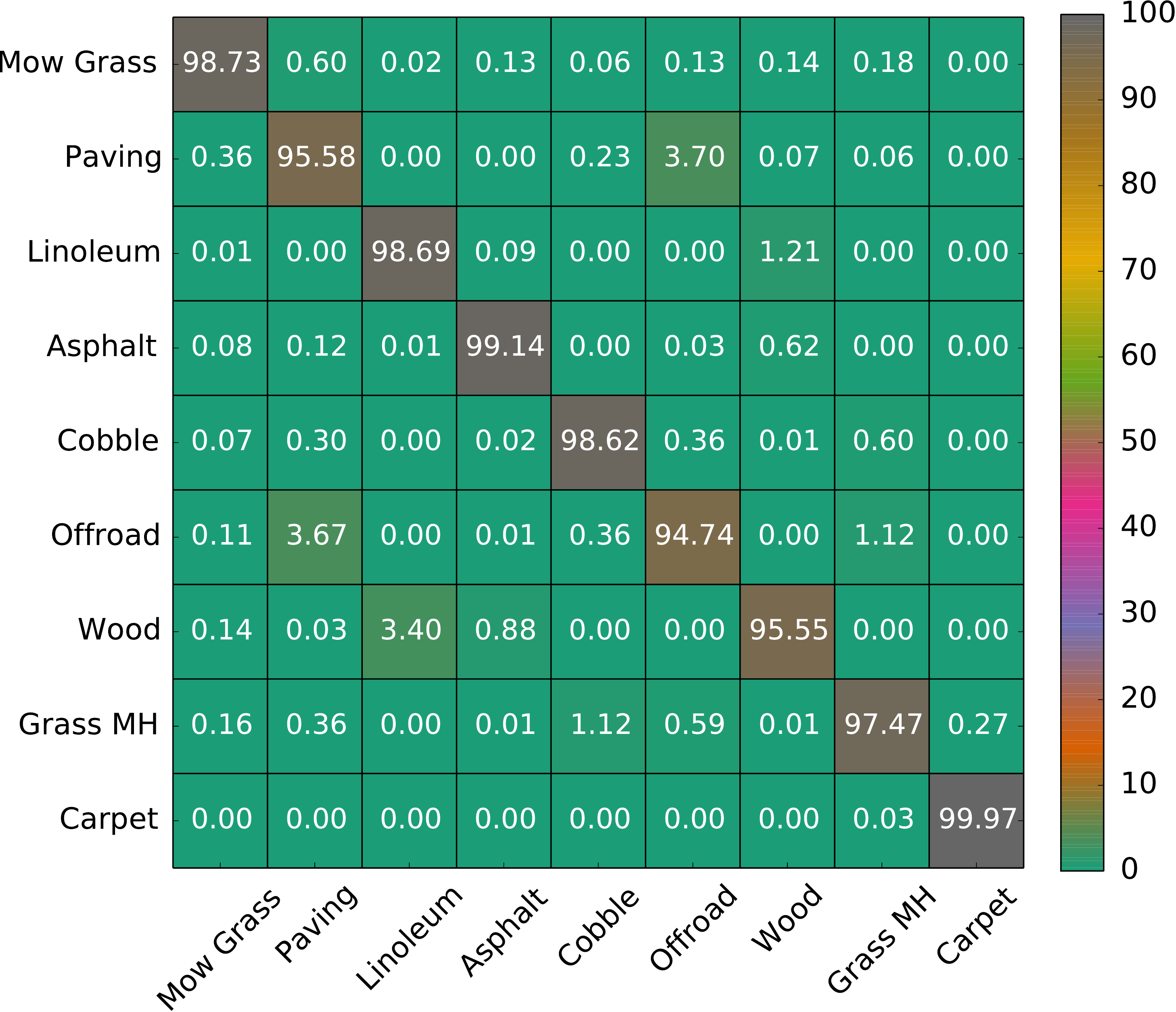}}
\hfill
\subfloat[Confusion matrix of our DCNN~ST model (M4) for a window length of five and clip length of $200\si{\milli\second}$\label{subfig-2:cm_200ms}]{%
\includegraphics[scale=0.4]{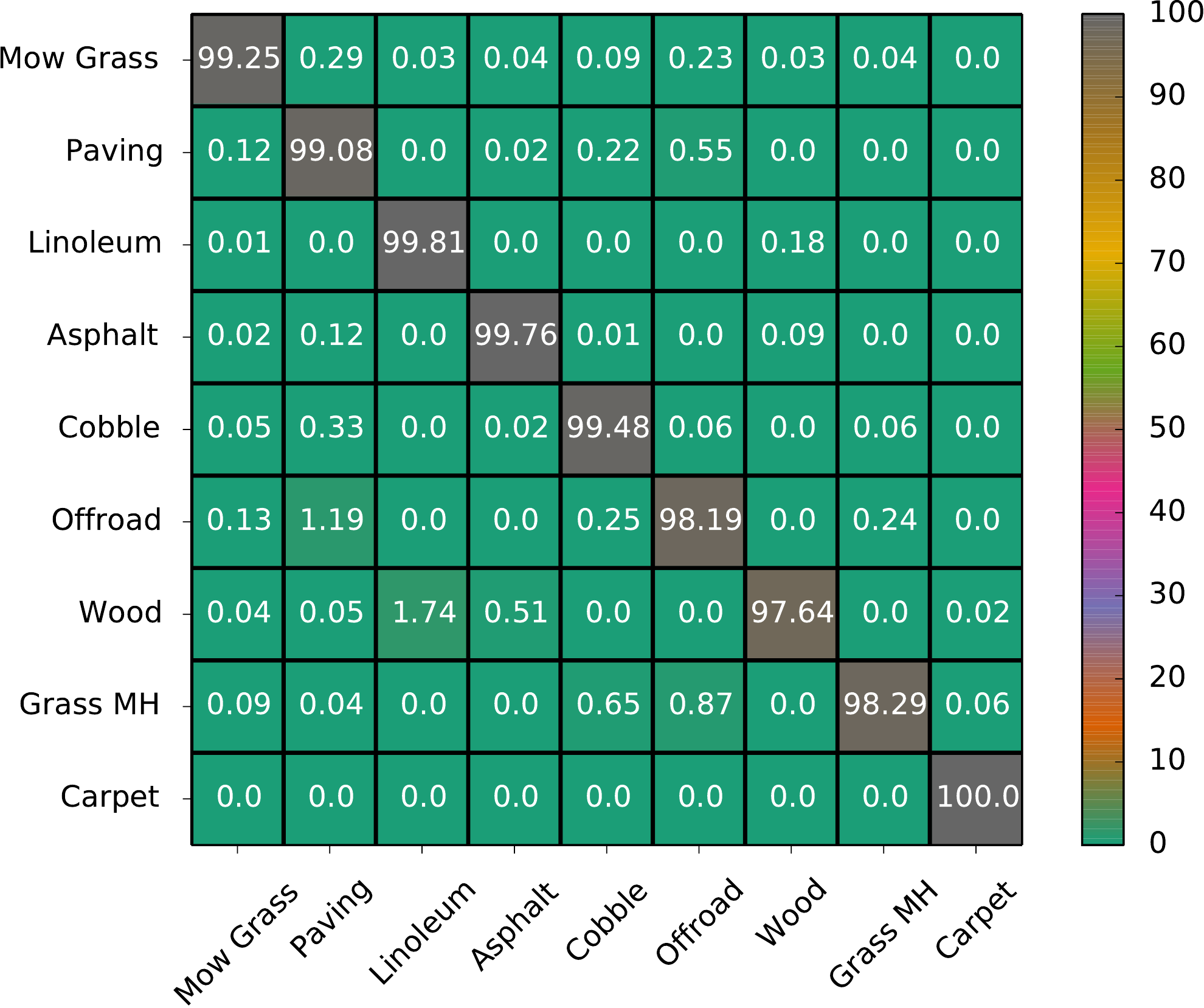}}
\hfill
\subfloat[Per-class recall of our DCNN model (M2) for a clip length of $300\si{\milli\second}$ \citep{valada2015isrr}\label{subfig-1:recall_300ms}\vfill]{%
\includegraphics[height=5.2cm,width=7.5cm]{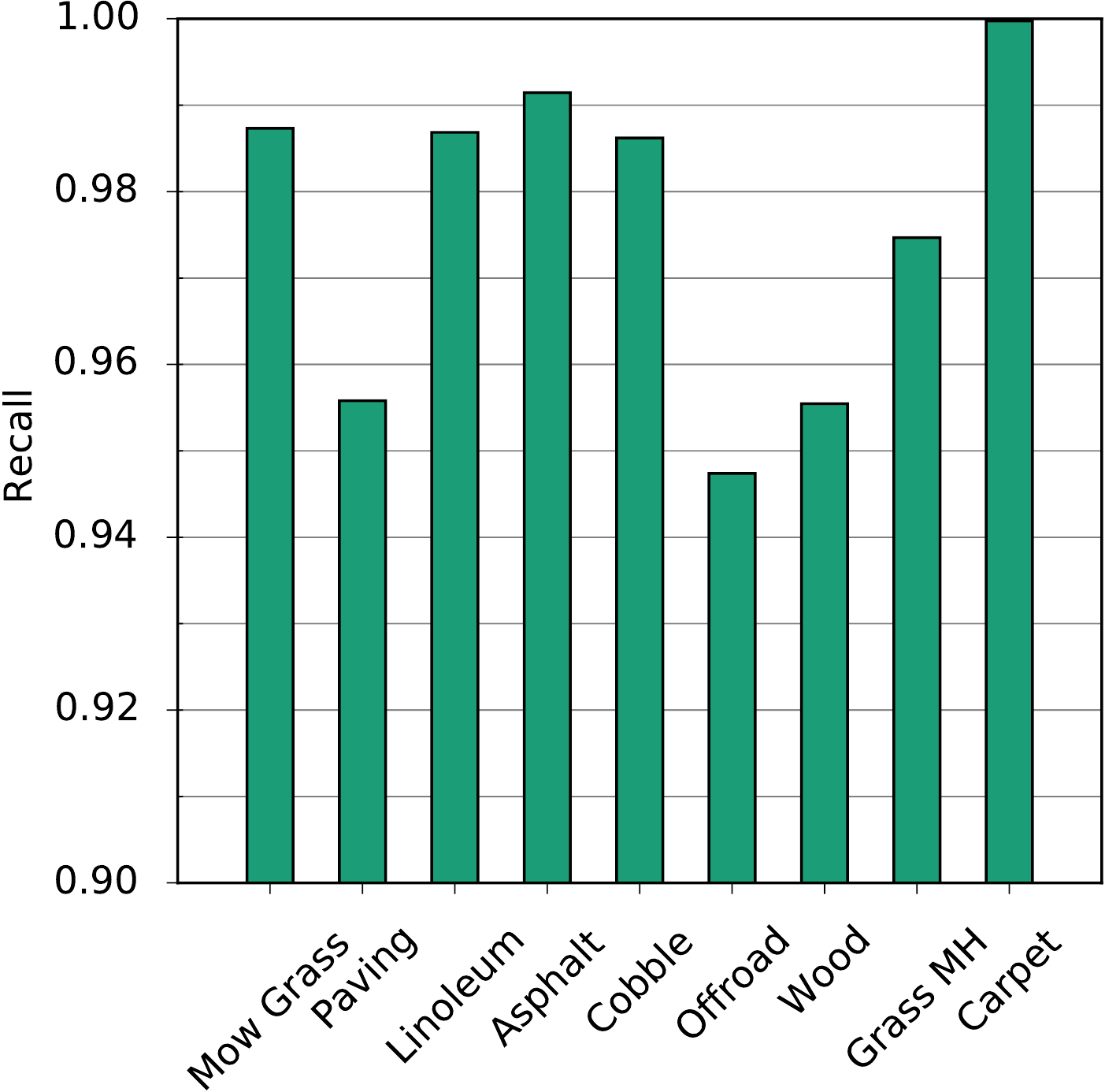}}
\hfill
\subfloat[Per-class recall of our DCNN~ST model (M4) for a window length of five and clip length of $200\si{\milli\second}$\label{subfig-2:recall_200ms}]{%
\includegraphics[scale=0.42]{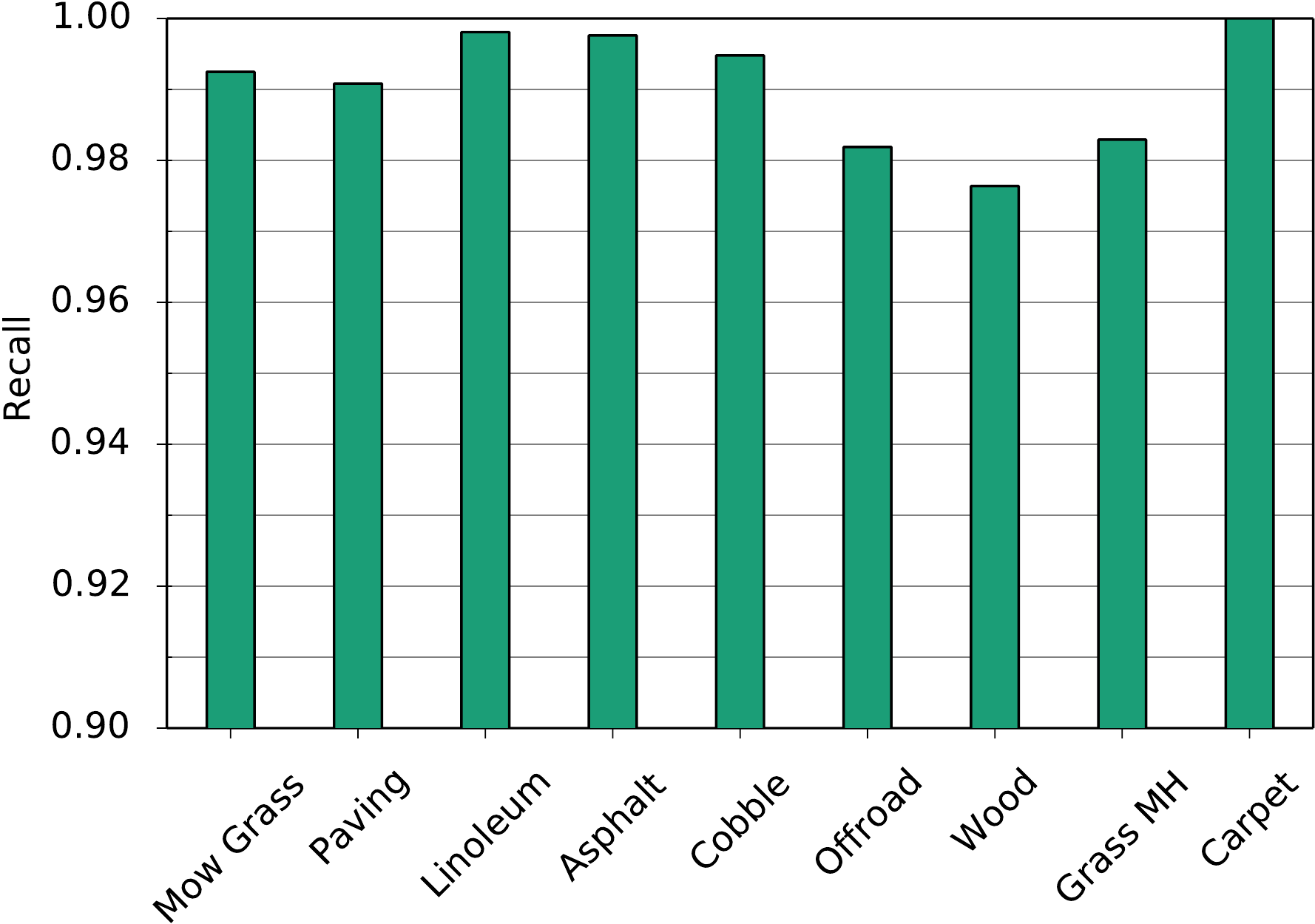}}
\caption{Comparison of our new DCNN~ST model with our DCNN model. Paving and Offroad classes have the highest decrease in the false positives and the Carpet class now demonstrates no false positives.}
\label{fig:percom}
\end{framed}
\end{figure*}

The results show that the accuracy does not increase just by considering a larger LSTM window length. In fact, we found the models to be increasingly difficult to train for large window lengths and it was difficult to ascertain when they actually converge even if we train them for twice the number of epochs as our base model. For a $300\si{\milli\second}$ clip, we obtained the best performance using a LSTM window size of three, above which the accuracy dropped significantly for a window size of four, and saturated for higher window lengths. As we experimented with lower clip sizes, we found the same pattern of obtaining the best performance for a certain window length and then a significant drop, followed by saturation for higher lengths. We also trained models with lower clip length than $200\si{\milli\second}$ and with higher LSTM window lengths but they did not yield a better performance than the reported models in Table~3.  The overall best performance for various LSTM window lengths and clip sizes was obtained using a clip length of $200\si{\milli\second}$ and with a window size of five.

Furthermore, we compared the performance of our DCNN model (M2) and our DCNN~ST model (M4) at varying audio clip lengths. Results from this comparison are shown in Table~4. It can be seen that our DCNN~ST model outperforms our base M2 model for all audio clip lengths between $200\si{\milli\second}$ and $2000\si{\milli\second}$. Furthermore, the DCNN~ST model also has faster classification rates for almost all clip lengths. The speed-up is achieved by computing gradients with respect to recurrent weights in a single matrix computation. Figure~8 shows the classification rates of both the models at varying audio clip lengths. Although the DCNN model has a faster classification rate for a clip length of $200\si{\milli\second}$, the accuracy is $7.73\%$ lower than that of the DCNN~ST model. Considering the clip length to classification time trade-off, we choose the $200\si{\milli\second}$ DCNN~ST model for the in-depth performance evaluation experiments described in the following sections.

\subsection{5.3. Performance Evaluation}

\begin{table*}
\centering
\caption{Influence of various ambient noises on the classification accuracy of our DCNN~ST model (M4). Noise samples are extracted from the DEMAND noise database. Noises in the White, Domestic and Street categories are the most damaging, whereas, noises in the Nature category are the least damaging.}
\label{tab:SNR}
\begin{tabular}{p{2cm} | p{1cm} | p{1cm}p{1cm}p{1cm}p{1cm}p{1cm} | p{1cm}}
\toprule
Noise & \multicolumn{7}{c}{SNR (dB)} \\
\cmidrule(lr){2-8}
 & Clean & 30 & 20 & 10 & 0 & -10 & mean \\
\midrule
White & 99.03 & 99.00 & 93.42 & 69.66 & 20.85 & 9.16 & 58.42  \\ 
Domestic & 99.03 & 98.63 & 82.84 & 55.24 & 25.38 & 9.16 & 54.25 \\
Nature & 99.03 & 99.03 & 99.02 & 98.99 & 98.63 & 90.27 & 97.19 \\
Office & 99.03 & 99.03 & 99.02 & 98.65 & 77.40 & 22.23 & 79.27 \\
Public & 99.03 & 99.02 & 99.01 & 98.17 & 73.01 & 12.99 & 76.44 \\
Street & 99.03 & 98.93 & 95.23 & 71.10 & 36.19 & 9.68 & 62.22 \\
Transportation & 99.03 & 99.03 & 98.99 & 97.48 & 57.02 & 28.65 & 76.23 \\
\midrule
mean & 99.03 & 98.95 & 95.36 & 84.18 & 60.87 & 26.02 &  \\
\bottomrule
\end{tabular}
\end{table*}

To further investigate the performance, we computed the confusion matrix, which gives us insights on the misclassifications between the classes. Figure~9(b) shows the confusion matrix of our best performing DCNN~ST model for a clip length of $200\si{\milli\second}$ and an LSTM window size of five. This corresponds to a substantial improvement in the misclassifications compared to our previous DCNN model (M2). Figure~9(a) shows the confusion matrix of our M2 model. Paving and Offroad, which had the highest misclassification rate previously, now has a $3.15\%$ decrease in the false positives. The second highest rate of false positives was between Wood and Linoleum. Our current DCNN~ST model demonstrates a $1.66\%$ decrease in the false positives between Wood and Linoleum. There is also a $0.88\%$ decrease in the misclassification between Offroad and Grass Medium-High. The misclassifications are primarily due to very similar spectral responses between the classes for short clip lengths.

The best performing classes were Carpet and Linoleum. In fact, the Carpet class demonstrated no false positives. All the classes other than Offroad, Wood and Grass Medium-High yielded accuracies in the high $99\%$ range. By incorporating LSTM units, our new proposed model outperforms our previous state-of-the-art model by $1.51\%$ and with a much smaller clip length. The per-class recall of this model is shown in Figure~9(d). This gives us insights on the ratio of correctly classified instances. The class with the lowest recall was Wood, whereas, the class with the highest recall was Carpet. The overall recall of the DCNN~ST model was $99.05\%$. 

\subsection{5.4. Robustness to Noise}

\begin{figure}
\begin{framed}
\centering
\includegraphics[width=\linewidth]{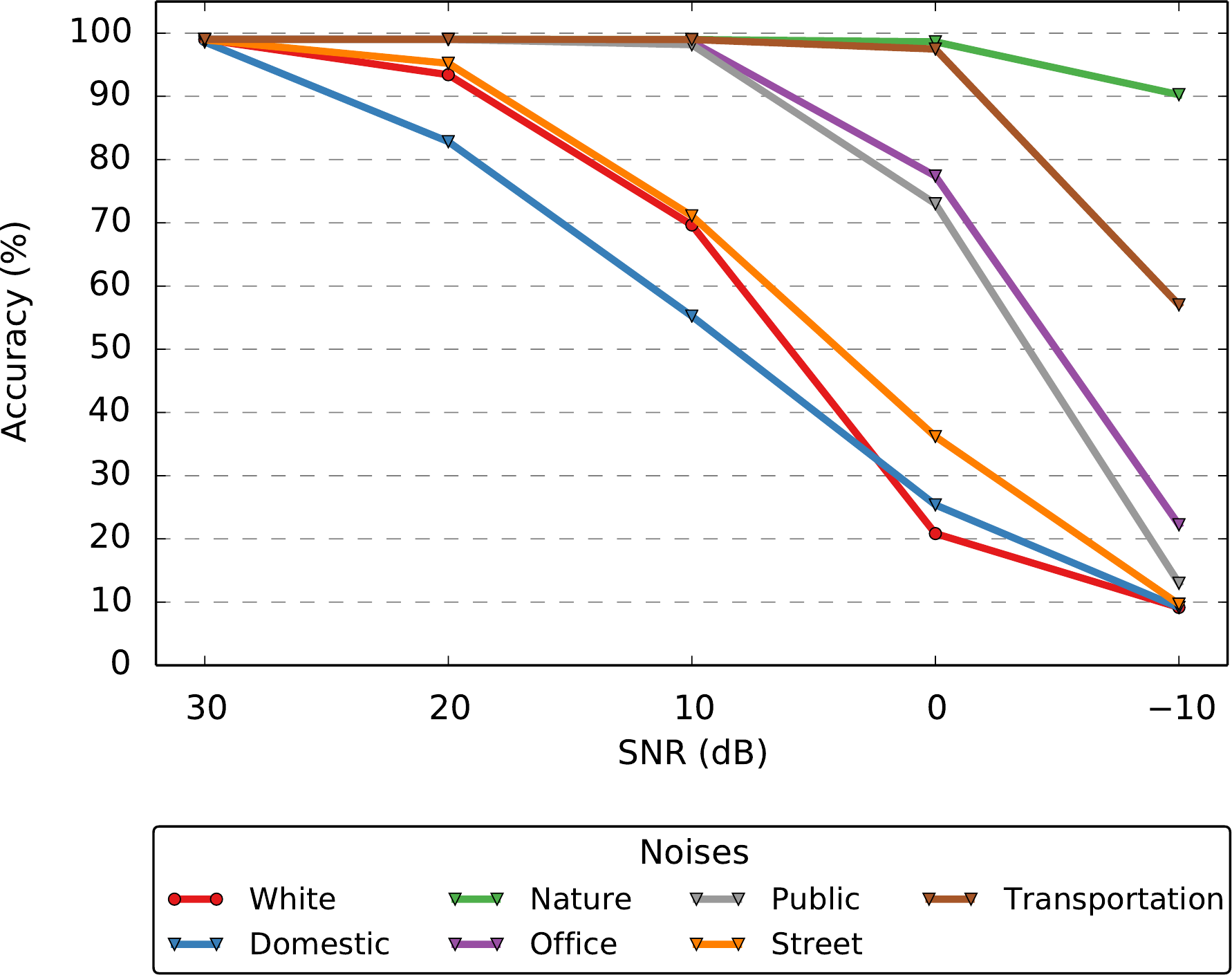}
\caption{Performance comparison of our DCNN~ST model when subject to different noises at varying SNRs.}
\label{fig:noisesComp}
\end{framed}
\end{figure}

In this section, we detail our experiments on the noise corrupted samples from the DEMAND noise database categorized into seven classes and at varying SNRs as described in Section~3.3. We first only added noise to our test set using methods described in \citep{loizou2007speech} and experimented on the model trained on clean noise-free signals. Table~5 shows detailed results from this experiment. We observe that the performance drastically decreases with decreasing SNRs. Moreover, different ambient noises affect the performance at varying degrees. Figure~10 shows the accuracy of our model for the various noises. It can be seen that for all the classes the models are fairly robust at SNRs until $20\decibel$, thereafter the performance drops rapidly for some classes such as noises from the Domestic and Street categories, as well as White noise. This can be attributed to the broadband of these noises, which corrupts signals in most frequencies, while the other types of noises are present only at specific frequencies. Hence, vehicle-terrain interaction signals consisting of high frequency components are relatively robust to noises such as in the category of Nature due to their presence mostly only in the low frequencies.

\begin{figure*}
\begin{framed}
\centering
\setlength{\tabcolsep}{0.5em}
\begin{tabular}{p{0.45\linewidth} p{0.45\linewidth}}
\includegraphics[width=\linewidth]{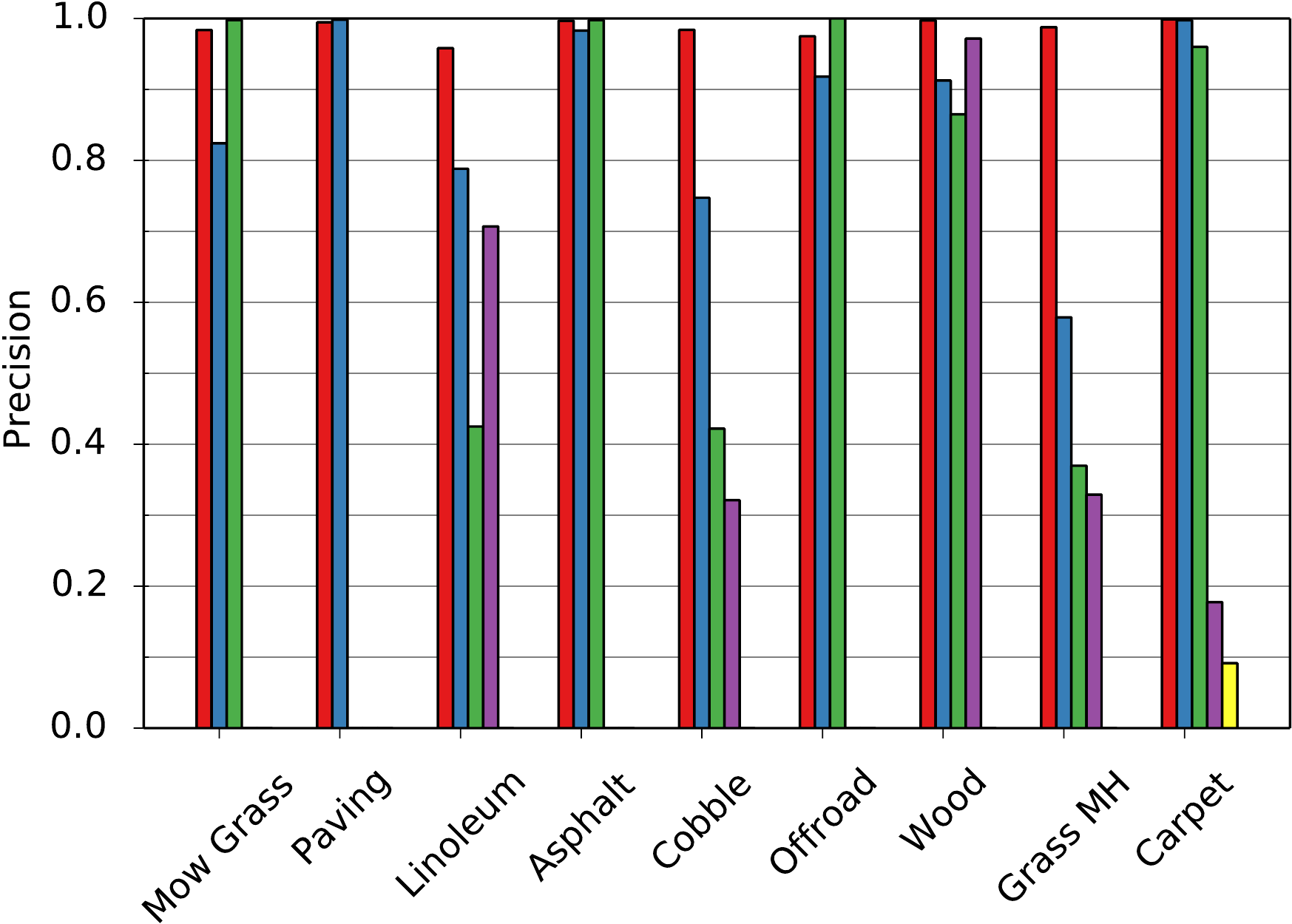} & 
\includegraphics[width=\linewidth]{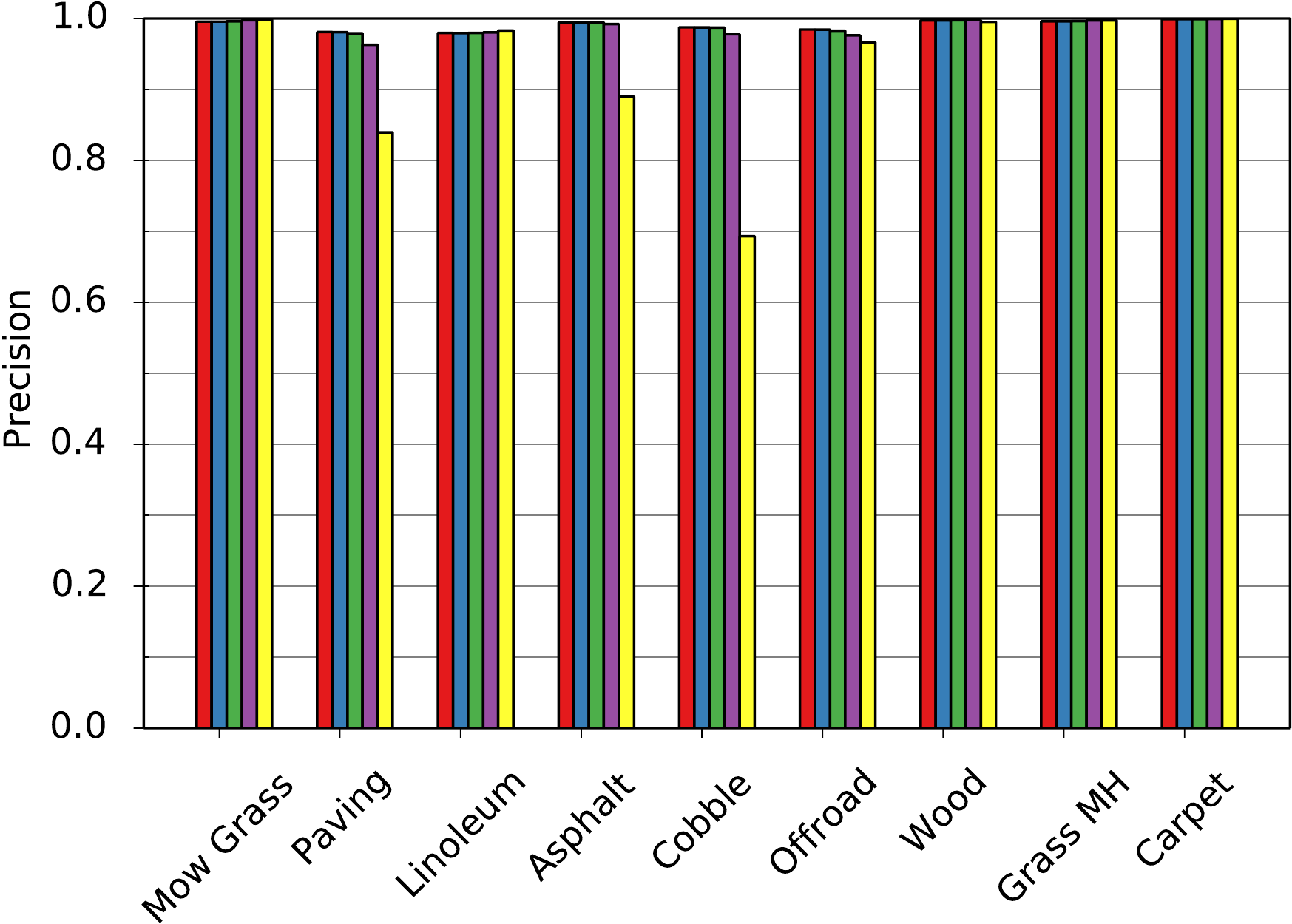} \\
{(a) Domestic Noise} & {(b) Nature Noise} \\
\\
\includegraphics[width=\linewidth]{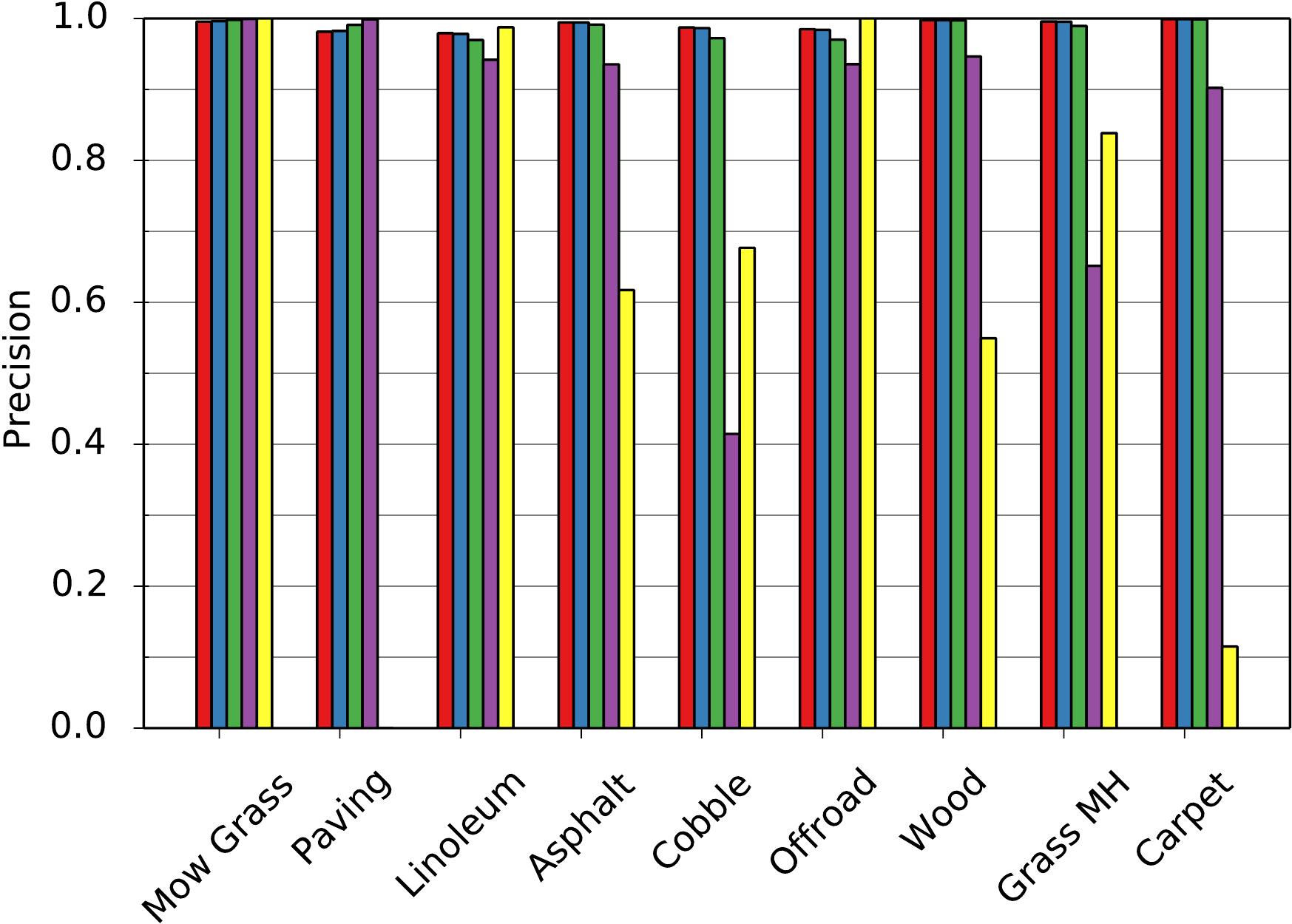} & 
\includegraphics[width=\linewidth]{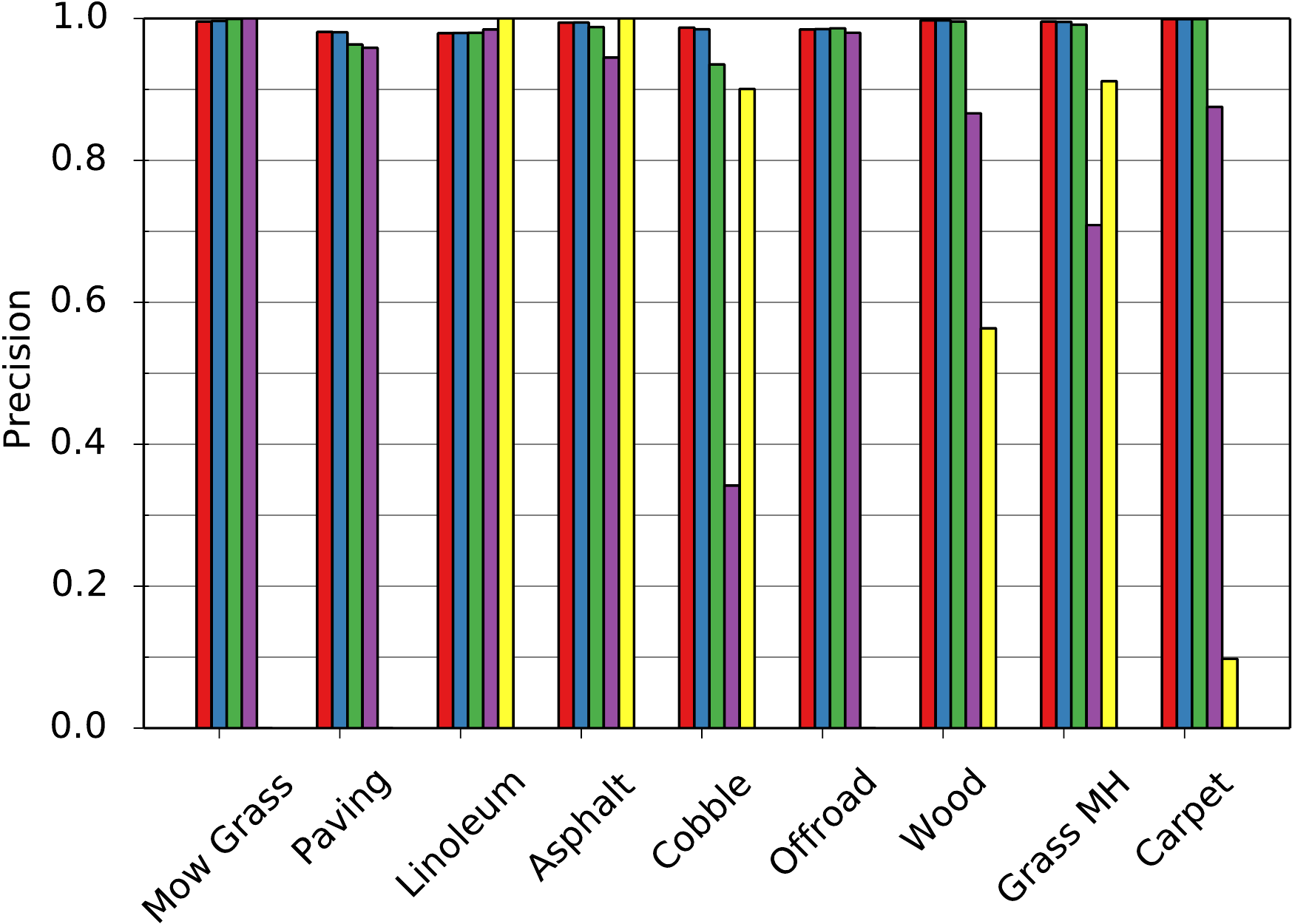} \\
{(c) Office Noise} & {(d) Public Noise} \\
\\
\includegraphics[width=\linewidth]{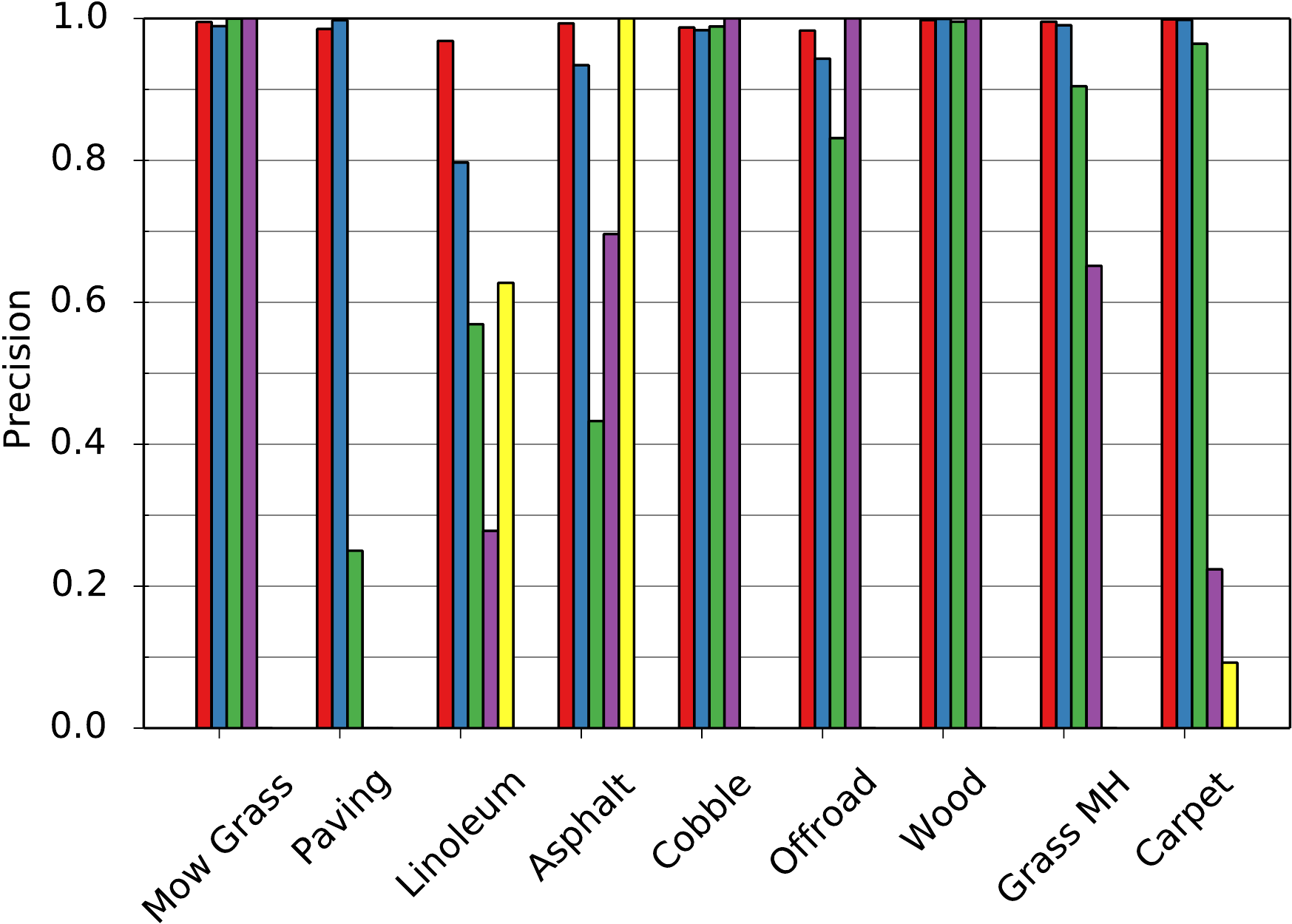} & 
\includegraphics[width=\linewidth]{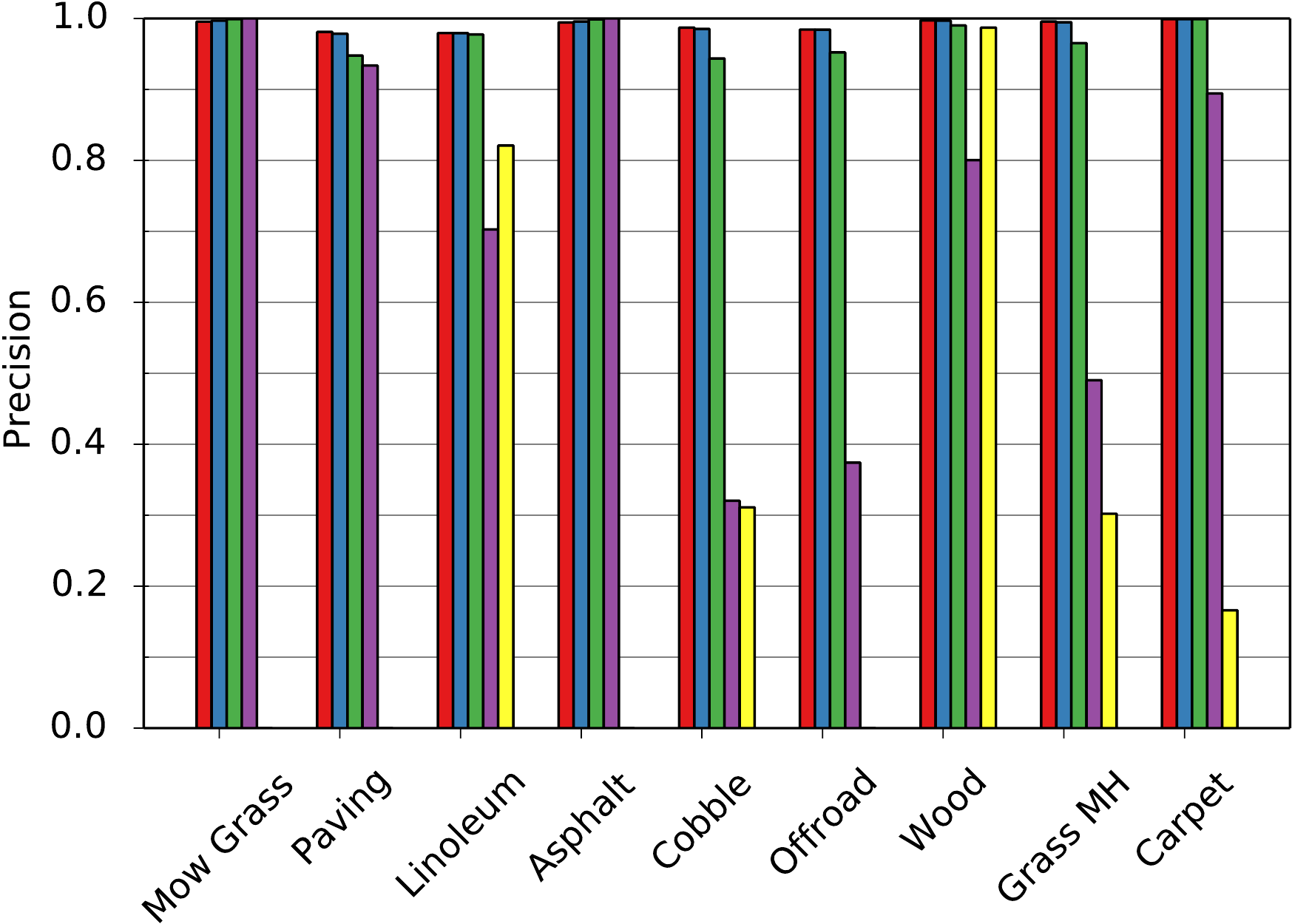} \\
{(e) Street Noise} & {(f) Transportation Noise} \\
\\
\multicolumn{2}{c}{\includegraphics[width=0.5\linewidth]{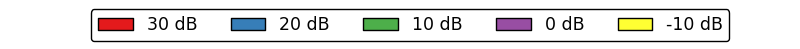}} \\
\end{tabular}
\caption{Per-class precision of our DCNN~ST model when subject to various noises from the DEMAND noise database at different SNRs. Terrains are still recognizable for ambient noises in the Nature category even for SNRs lower than $0\decibel$.}
\end{framed}
\end{figure*}

\begin{table*}
\centering
\caption{Influence of various ambient noises on the classification accuracy of our noise-aware DCNN~ST model. There is a substantial increase in the robustness in all the noise categories.}
\label{tab:SNR}
\begin{tabular}{p{2cm} | p{1cm} | p{1cm}p{1cm}p{1cm}p{1cm}p{1cm} | p{1cm}}
\toprule
Noise & \multicolumn{7}{c}{SNR (dB)} \\
\cmidrule(lr){2-8}
 & Clean & 30 & 20 & 10 & 0 & -10 & mean \\
\midrule
White & 99.72 & 99.68 & 98.77 & 97.63 & 97.11 & 96.38 & 97.91 \\ 
Domestic & 99.72 & 99.66 & 98.69 & 97.87 & 97.04 & 96.97 & 98.05 \\
Nature & 99.72 & 99.71 & 99.69 & 99.51 & 99.08 & 97.28 & 99.05 \\
Office & 99.72 & 99.72 & 99.65 & 98.70 & 98.61 & 96.00 & 98.53 \\
Public & 99.72 & 99.70 & 99.67 & 98.71 & 98.49 & 97.63 & 98.84 \\
Street & 99.72 & 99.68 & 98.73 & 97.90 & 97.81 & 96.36 & 98.1 \\
Transportation & 99.72 & 99.69 & 98.77 & 98.62 & 97.65 & 97.47 & 98.44 \\
\midrule
mean & 99.72 & 99.69 & 99.13 & 98.42 & 97.97 & 96.87 &  \\
\bottomrule
\end{tabular}
\end{table*}

Ambient noises in the domestic category have the most damaging effect, where the mean accuracy at SNRs from $30\decibel$ to $-10\decibel$ was $61.71\%$. It can be seen in Figure~10 that the model has a linear drop in accuracy for decreasing SNRs, for noises from the Domestic category. The second most damaging noises were from the Street category, achieving a mean accuracy of $68.36\%$. The model demonstrates remarkable robustness to ambient noises from the Nature category, for which the model yielded an average accuracy of $97.49\%$, even for such low SNRs. This is primarily because noises in the Nature category have very distinct spectral patterns that have dominant structures, unlike vehicle-terrain interaction signals, hence the model is inherently robust to these corruption patterns.

\begin{figure}
\begin{framed}
\centering
\includegraphics[width=\linewidth]{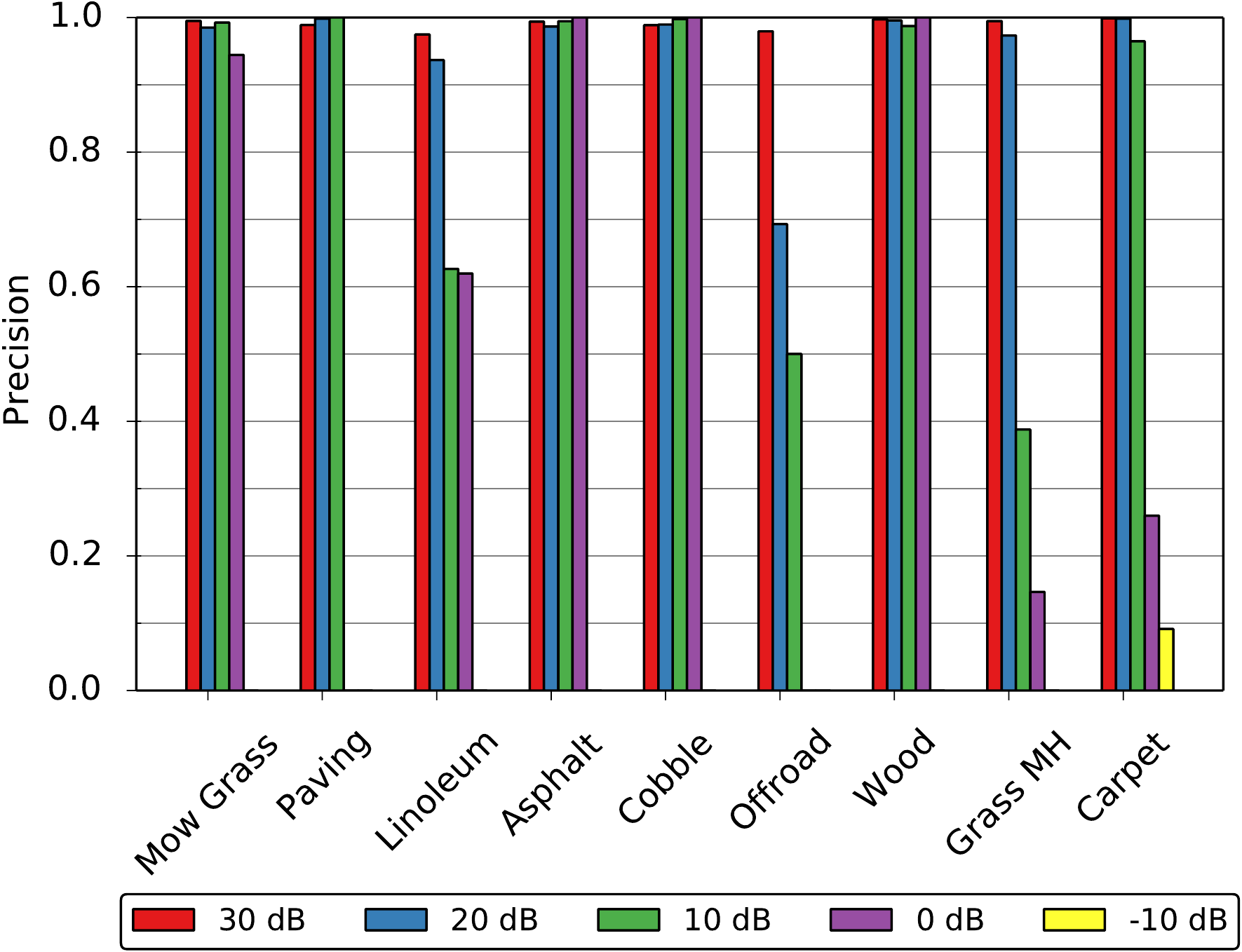}
\caption{Per-class precision of our DCNN~ST model when subject to different levels of white Gaussian noise. The levels mentioned in the legend are SNRs.}
\label{fig:whiteNoised}
\end{framed}
\end{figure}

\begin{figure*}
\begin{framed}
\centering
\setlength{\tabcolsep}{0.3em}
\begin{tabular}{p{0.3\linewidth} p{0.3\linewidth} p{0.3\linewidth}}
\includegraphics[width=\linewidth]{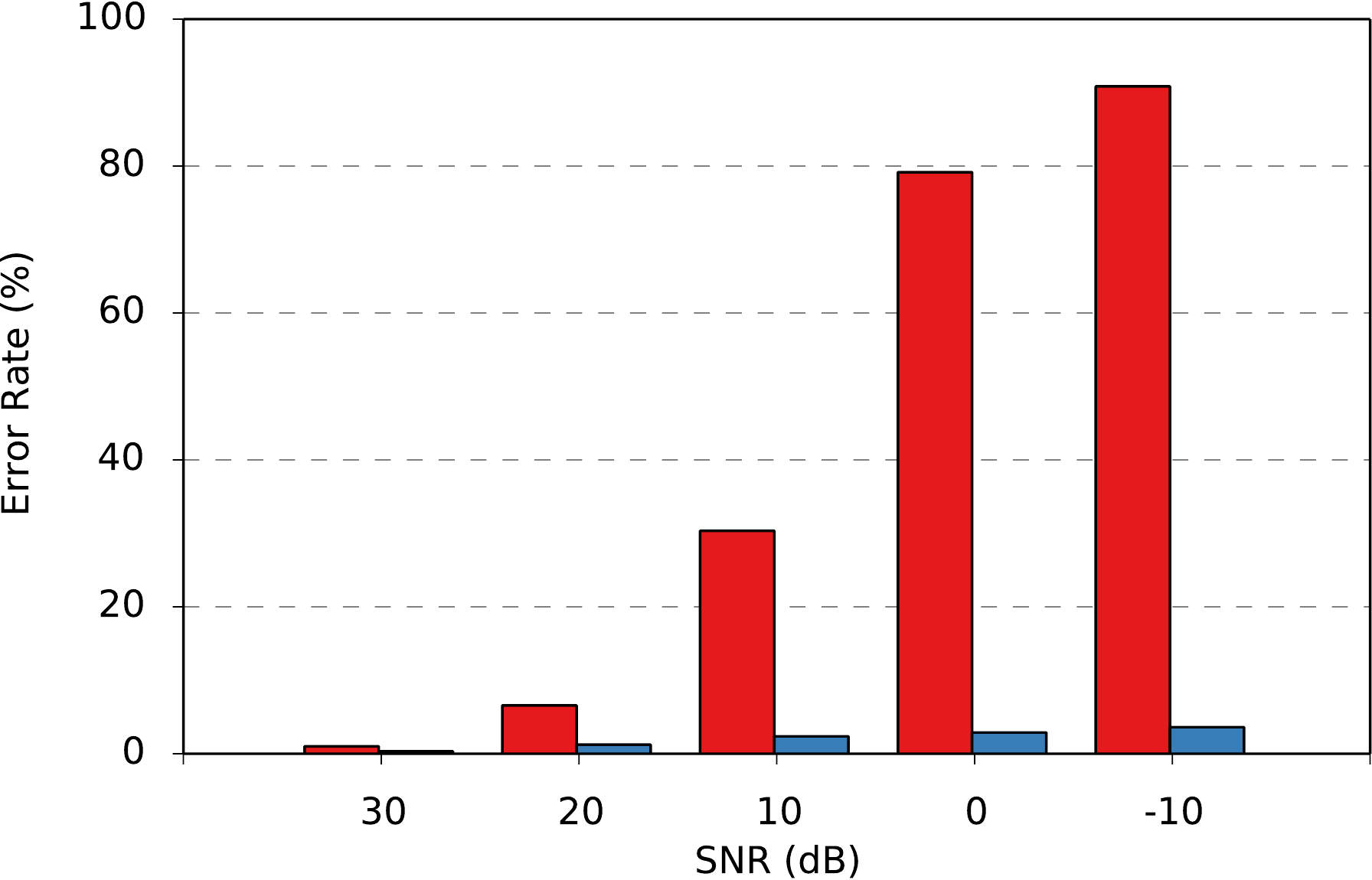} & 
\includegraphics[width=\linewidth]{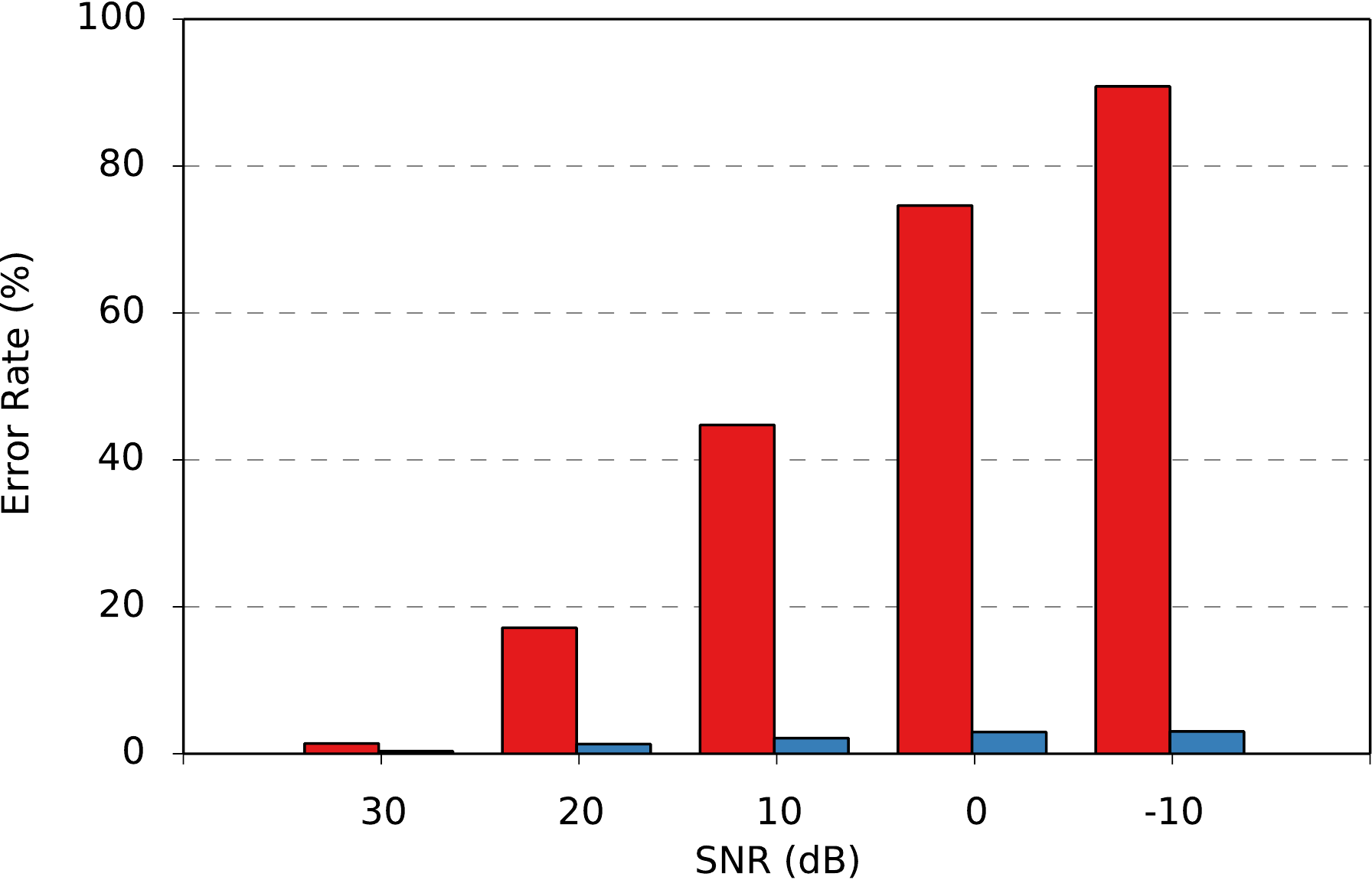} &
\includegraphics[width=\linewidth]{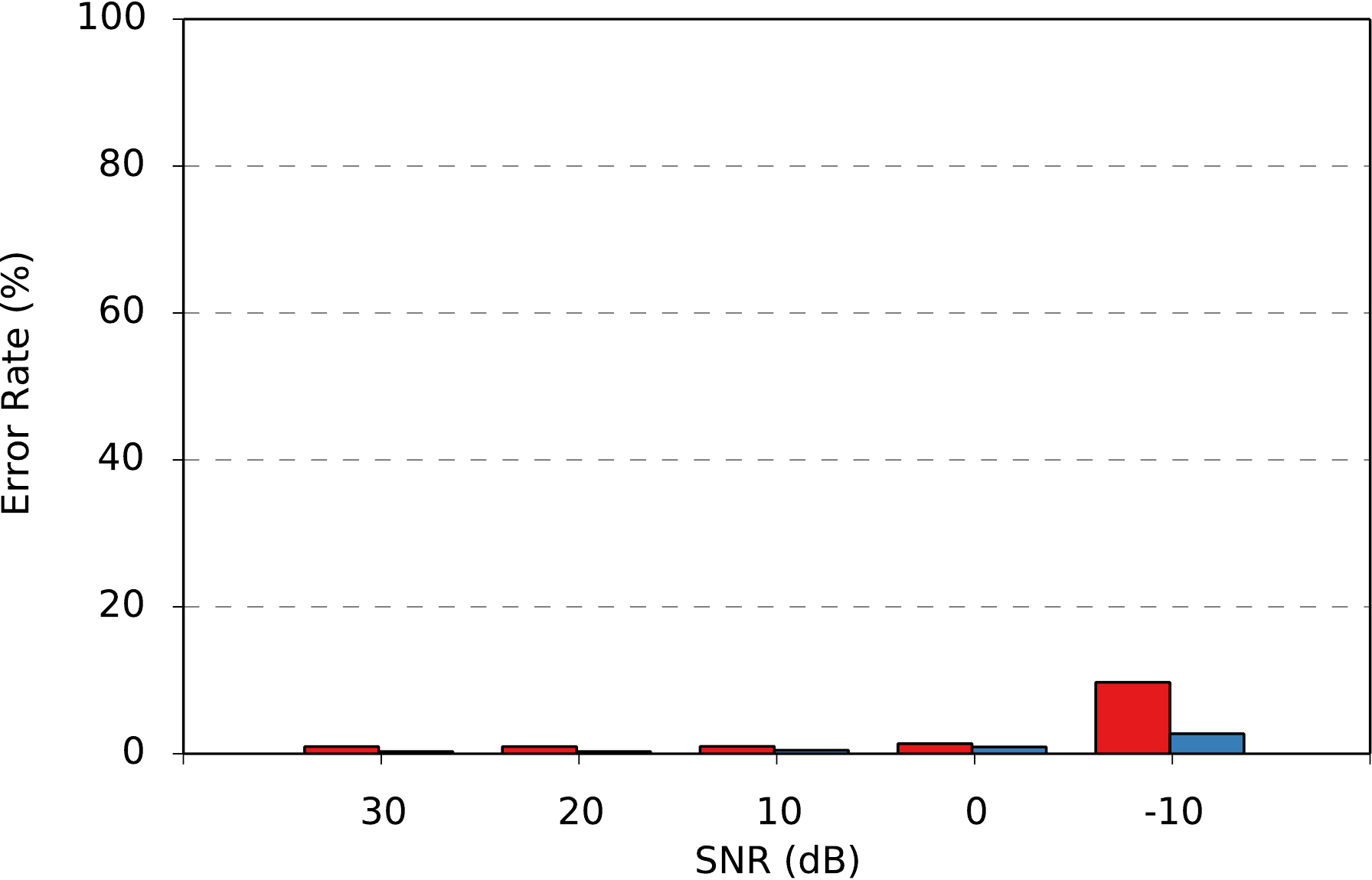} \\
{(a) White Noise} & {(b) Domestic Noise} & {(c) Nature Noise} \\
\\
\includegraphics[width=\linewidth]{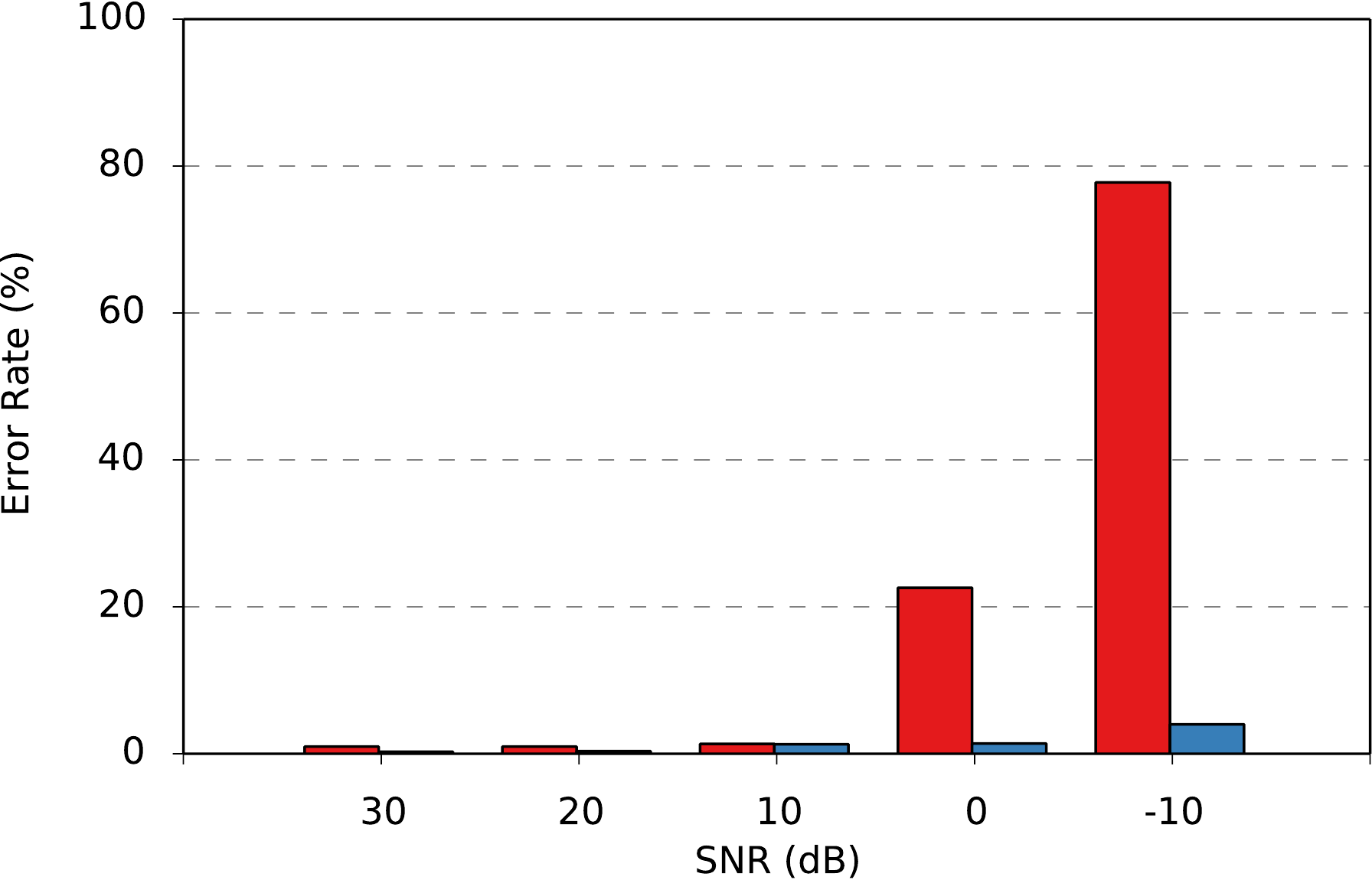} &
\includegraphics[width=\linewidth]{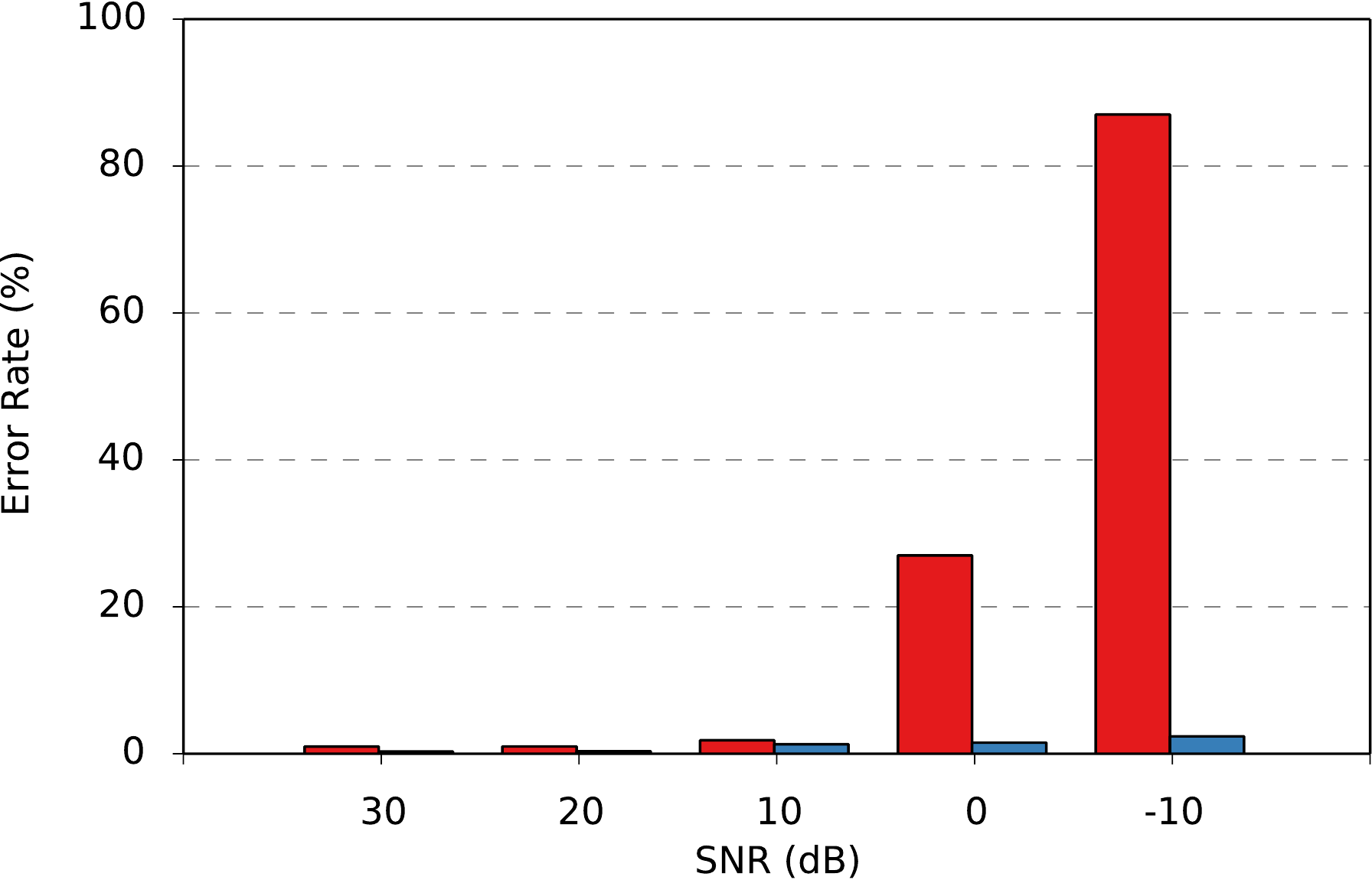} & 
\includegraphics[width=\linewidth]{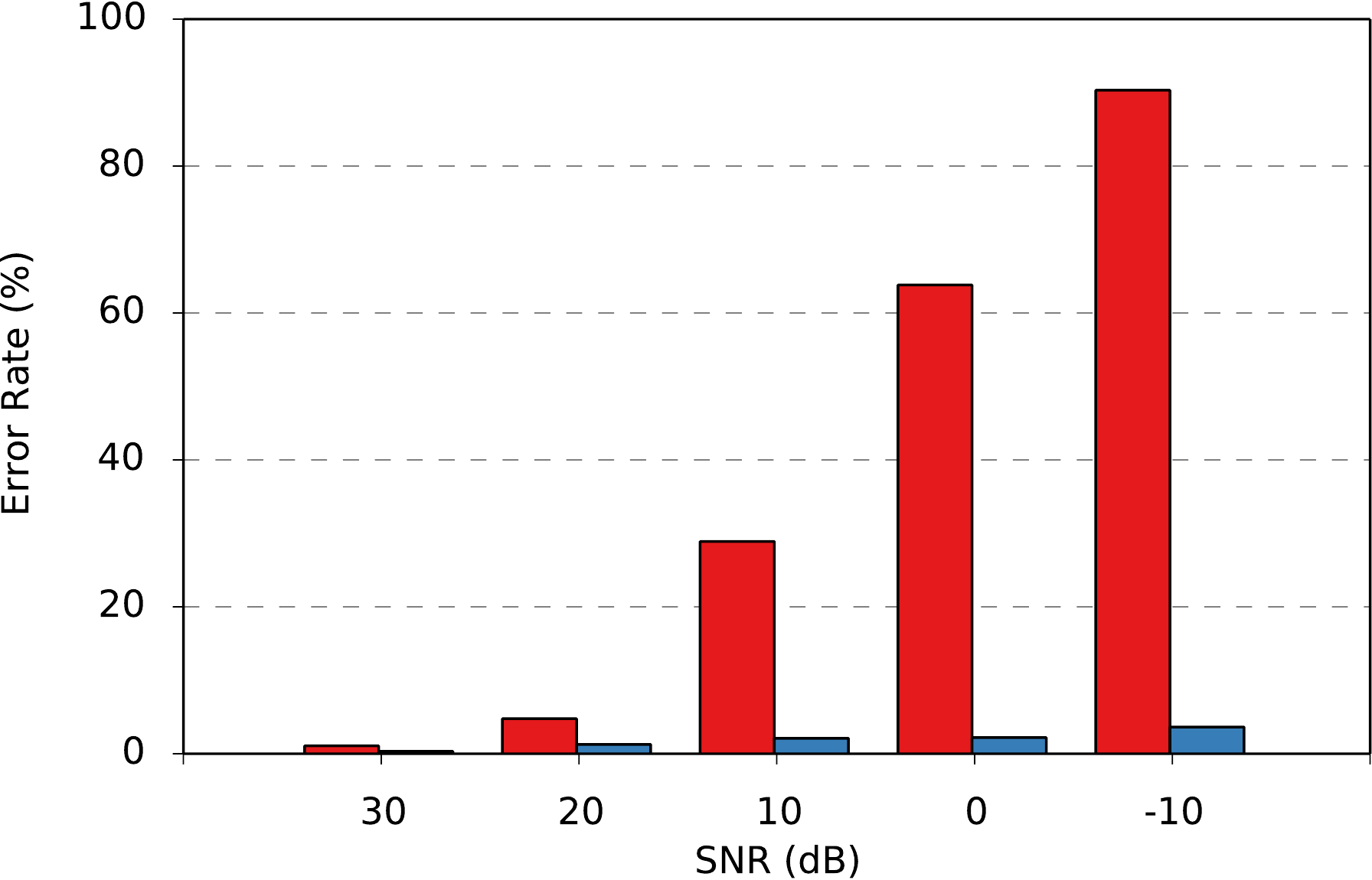} \\
{(d) Office Noise} & {e) Public Noise} & {(f) Street Noise} \\
\\
& \includegraphics[width=\linewidth]{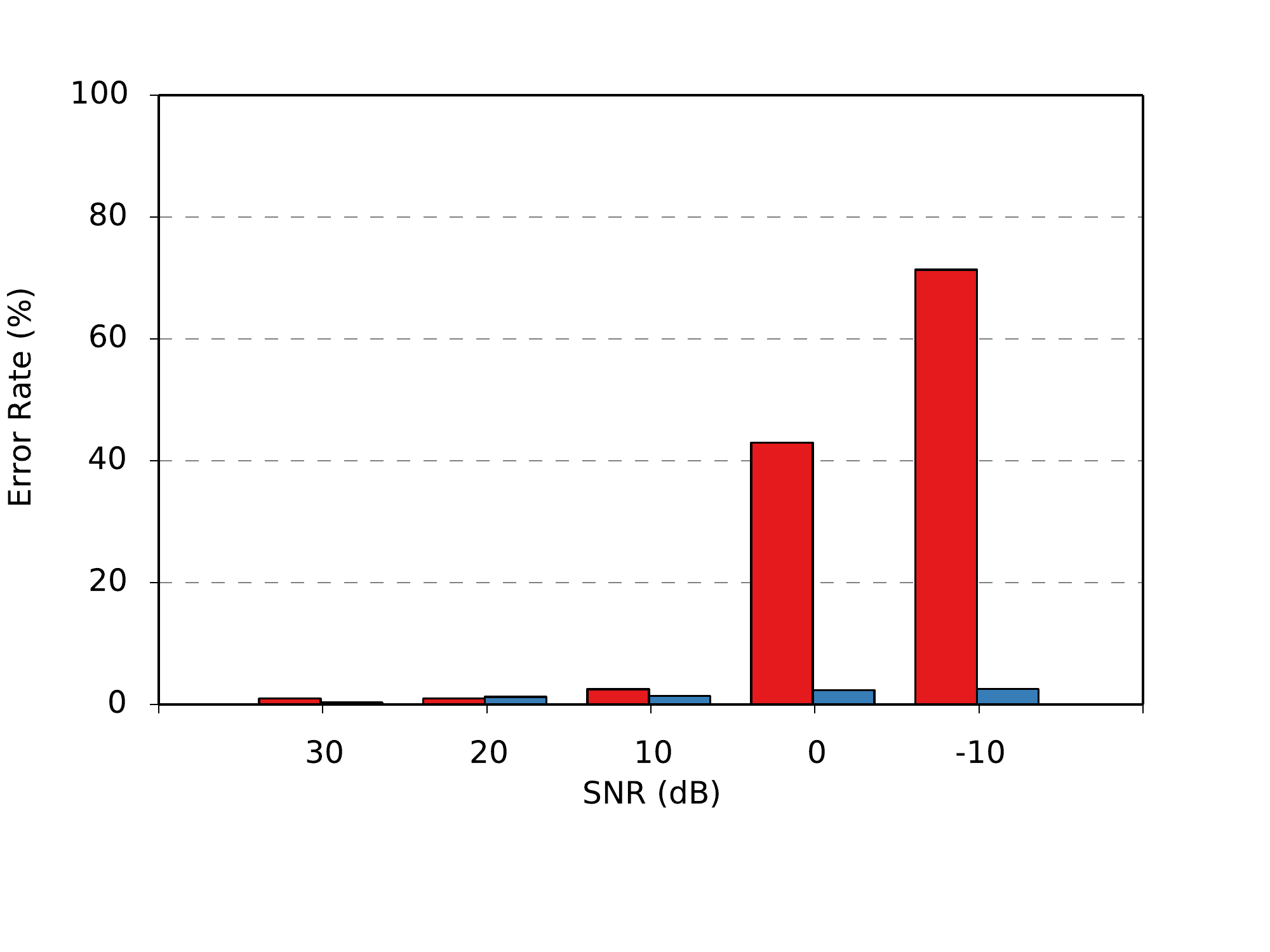} & \\
& {g) Transportation Noise} & \\
& \includegraphics[width=\linewidth]{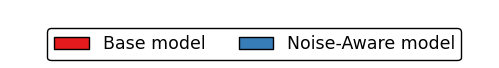} & \\
\end{tabular}
\caption{Comparison of classification error rates of our base DCNN~ST model and our noise-aware DCNN~ST model. Our noise-aware DCNN~ST model demonstrates much lower error than our base DCNN~ST model.}
\end{framed}
\end{figure*}

It is also of interest to know how each of these noises influence the classification of specific terrains. Therefore we computed the per-class precision for each of the noise categories. Results from this experiment are shown in Figure 11 for ambient noises from the DEMAND noise database and in Figure 12 for White noise. The Carpet class is recognizable for all the ambient noises and at all SNRs. The Paving class is the most affected by noise corruption, followed by Offroad and Asphalt. Interestingly, classification of indoor terrains such as Linoleum, Wood and Carpet are less affected by indoor noises than outdoor terrains. Although the converse does not appear to be true, where the outdoor terrains are less affected by outdoor noises. This is due to the reason that, generally, outdoor ambient noises have continuous noise corruption for long periods, whereas indoor ambient noises are usually short impacts. A curious phenomenon that can be observed in certain terrain classes for noise corruption with SNRs of $-10\decibel$ with ambient noises such as Office, Public, Street and Transportation, is the sudden increase in precision compared to higher SNRs. This is primarily due to the fact that the spectral responses of these noises blur the inherently noisy lower frequency components of the vehicle-terrain interaction signals which enhances the ability of the network to classify these terrains by focusing on the more distinct higher frequency responses.

White noise on the other hand severely affects the classification ability below $0\decibel$. Terrain classes such as Mowed-grass, Asphalt, Cobble and Wood are robust to noise corruption upto $0\decibel$. Whereas, classes such as Linoleum, Offroad, Grass Medium-high and Carpet have an exponential increase in classification error for decreasing SNRs. Paving and Offroad terrains are not recognizable for SNRs greater than $10\decibel$. These experiments demonstrate the need for noise robustness in order to operate in high ambient noise environments.

As described in Section~3.3, we performed noise-aware training by injecting ambient noises from the DEMAND noise database at varying SNRs. The noise injection helps the network learn  probable noise corruption patterns that may occur in real-world environments. Results from this experiment are shown in Table~6. It can be observed that there is a substantial increase in classification accuracy for SNRs lower than $10\decibel$. Even for very low SNRs, the model demonstrates state-of-the-art performance. The noise-aware model shows an improved performance even for pure signals without noise corruption. A comparison of the error rates of our base DCNN~ST model and our noise-aware DCNN~ST model for each of the noise categories in the DEMAND noise database is shown in Figure~12. Our base model demonstrated varying patterns of error change for increasing SNRs, whereas our noise-aware model shows a very gradual exponential increase for all the noise categories. This shows that noise-aware training enables the network to learn the general distribution of noise corruption for all the SNRs and for varying types of noises as well. 

For the second part of the robustness experiments, we tested our model on our mobile phone microphone dataset. This dataset was collected at a different location than our main dataset on which we trained our models. Each of the samples in this dataset was tagged with a GPS location for visualizing the trajectory in maps and to correlate to the terrain. This dataset contains about $2.5\hour$ of audio. Unlike the shotgun microphone that we used for collecting our main dataset, mobile phones have a condenser microphone that collects sound from every direction, thereby also collecting considerable amount of ambient noise. The main purpose of testing on this dataset was threefold:
\begin{enumerate*}
\item to verify the adaptability of the model to a new environment;
\item to quantify the models performance with a new hardware; and 
\item to quantify the performance in the presence of substantial amount of real-world ambient noise.
\end{enumerate*}

\begin{figure*}
\begin{framed}
\centering
\subfloat[Confusion matrix of our DCNN model (M2) for a clip length of $300\si{\milli\second}$ \citep{valada2015isrr}\label{subfig-1:cm_phone_300ms}\vfill]{%
\includegraphics[scale=0.26]{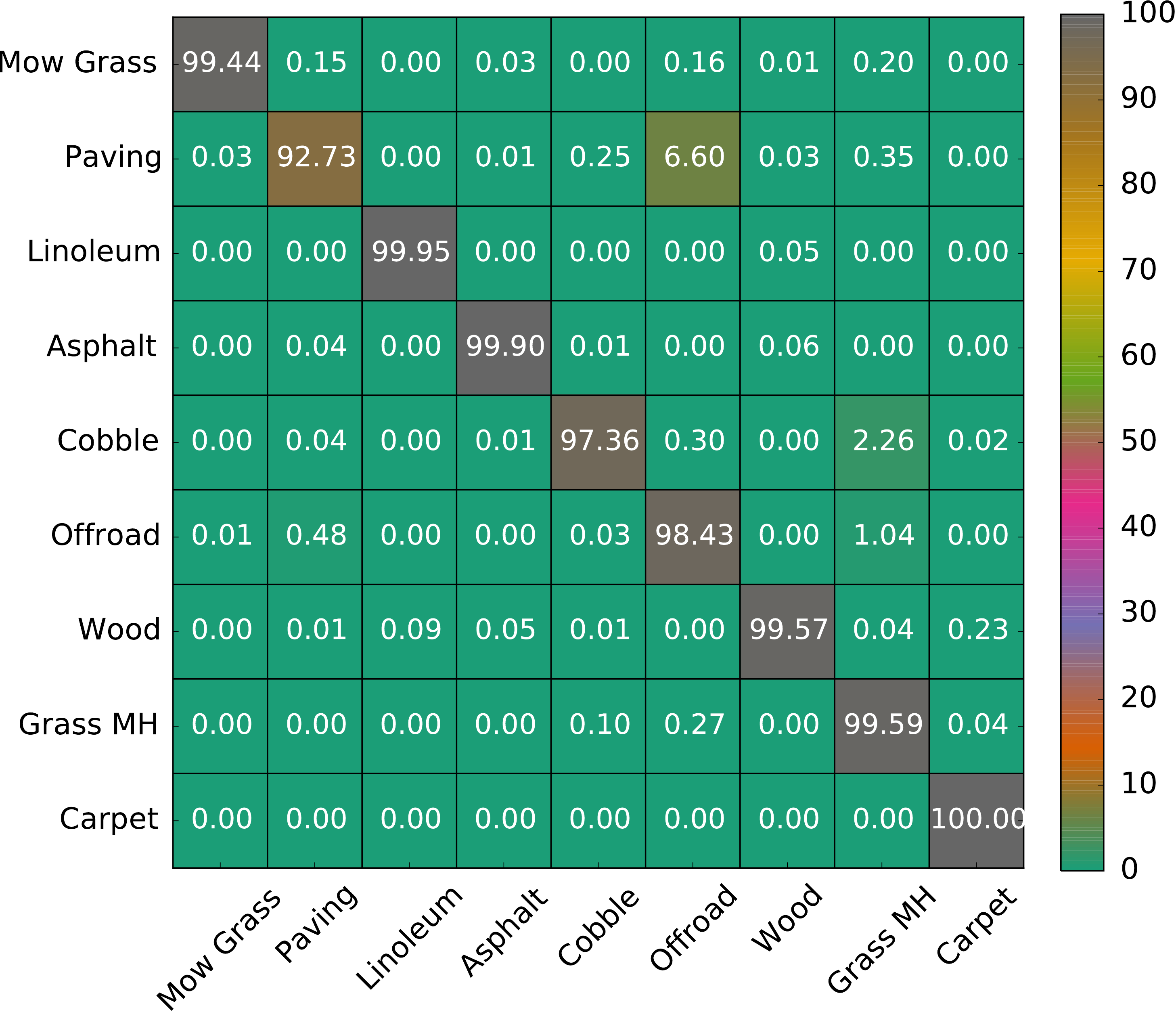}}
\hfill
\subfloat[Confusion matrix of our DCNN~ST model (M4) for a window length of five and clip length of $200\si{\milli\second}$\label{subfig-2:cm_phone_200ms}]{%
\includegraphics[scale=0.38]{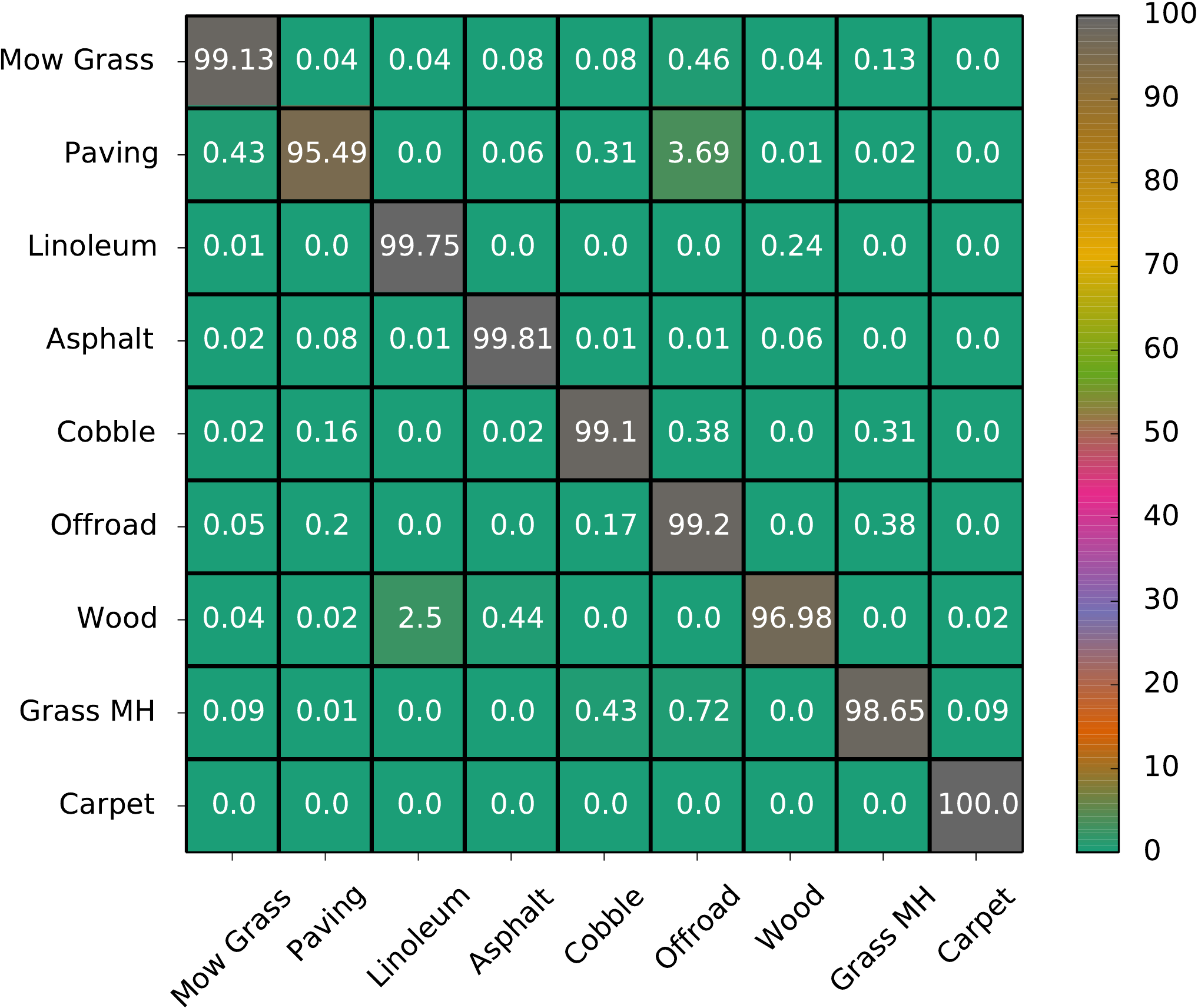}}
\caption{Comparison of our new DCNN~ST model with our DCNN model on the mobile phone dataset. There is a significant decrease in misclassifications between Paving and Ofroad classes, as well as Cobble and Grass Medium-High.}
\label{fig:percomp_phone}
\end{framed}
\end{figure*}

Figure~14 shows an example trajectory that the robot traversed during experimentation. The figure also shows the variation in speed $(0-2\meter\per\second)$ along the path. The speed at which the robot traversed with along the path is indicated with red lines. Thicker the red lines, slower is the speed. Our model achieves an accuracy of $98.62\%$ on this dataset. Figure~15 shows the confusion matrices of our DCNN model and DCNN~ST model tested on the mobile phone microphone dataset. The classes that show the largest misclasifications using our DCNN~ST model are Paving, Wood and Linoleum. There is about $3\%$ decrease in misclassifications between Paving and Offroad in our DCNN~ST model, while compared to the DCNN model. There is also an equivalent decrease in misclassification between the Cobble and Offroad classes. Compared to our DCNN model, the DCNN~ST model shows an increase in misclassification between Wood and Linoleum, however, the overall accuracy of the DCNN~ST model is still higher than our previous DCNN model. These experiments have demonstrated the utility of noise-aware training, making it a necessary step for robust terrain classification in noisy real-world environments.

\begin{figure}
\begin{framed}
\centering
\includegraphics[width=\linewidth]{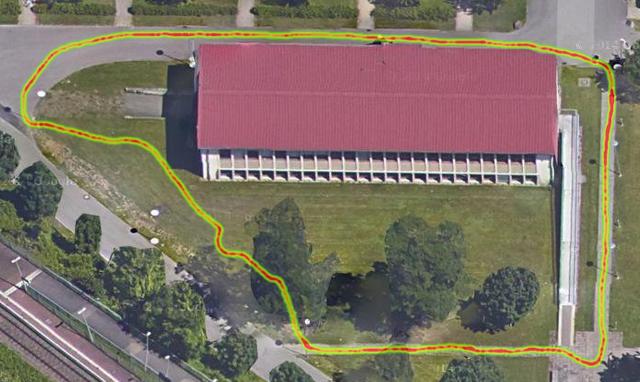}
\caption{Map showing one of the trajectories that the robot followed during the classification experiments using a mobile phone microphone. Variation in speed along the path is shown in red. Thicker red lines denote slower speed.}
\label{fig:mapPhoneExp}
\end{framed}
\end{figure}

\section{6. Conclusion}
\label{conclusion}

In this work, we introduced a robust proprioceptive terrain classification system based on recurrent convolutional neural networks, which uses only sound from vehicle-terrain interactions to classify a wide range of indoor and outdoor terrains. The performance of our models surpass several baseline approaches, achieving state-of-the-art results in proprioceptive terrain classification. Our proposed model is both spatially and temporally deep, and our results demonstrate that learning temporal dynamics can improve classification than when learning only in the spatial domain. We investigated the influence of various hyperparameters on the performance of our network and tuned them to obtain the best classification rate to accuracy ratio. Our implementation operates on a $200\milli\second$ window at $81.7\hertz$ and achieves an overall accuracy of $99.03\%$. 

Additionally, we thoroughly evaluated the robustness of our models to various extreme ambient noises and introduced a noise-aware training scheme that injects random noise into the training data that increases the overall robustness to a great extent. Our noise-aware model achieves an improved accuracy of $99.72\%$ and the extensive experiments demonstrate the capability of our model to adapt to real-world environments with different ambient noises. We also presented empirical evaluations with an inexpensive low-quality microphone that shows the hardware independence and robustness of our approach.

\begin{funding}
This work has been partly supported by the European Commission under the grant numbers ERC-AGPE7-267686-LifeNav and FP7-610603-EUROPA2.
\end{funding}

\bibliographystyle{abbrvnat}
\bibliography{references}

\begin{thebibliography}{41}
\providecommand{\natexlab}[1]{#1}
\providecommand{\url}[1]{\texttt{#1}}
\expandafter\ifx\csname urlstyle\endcsname\relax
  \providecommand{\doi}[1]{doi: #1}\else
  \providecommand{\doi}{doi: \begingroup \urlstyle{rm}\Url}\fi

\bibitem[Best et~al.(2013)Best, Moghadam, Kottege, and
  Kleeman]{best2013terrain}
G.~Best, P.~Moghadam, N.~Kottege, and L.~Kleeman.
\newblock Terrain classification using a hexapod robot.
\newblock In \emph{Australasian Conference on Robotics and Automation}, 2013.

\bibitem[Boersma and Weenink(2013)]{praat2013}
P.~Boersma and D.~Weenink.
\newblock Praat: doing phonetics by computer [computer program].
\newblock In \emph{Version 5.3.51, retrieved 2 June 2013 from
  http://www.praat.org/}, 2013.

\bibitem[Brooks and Iagnemma(2005)]{brooks2005ta}
C.~A. Brooks and K.~Iagnemma.
\newblock Vibration-based terrain classification for planetary exploration
  rovers.
\newblock \emph{IEEE Transactions on Robotics}, 21\penalty0 (6):\penalty0
  1185--1191, Dec 2005.

\bibitem[Brooks and Iagnemma(2007)]{brooks2007ac}
C.~A. Brooks and K.~D. Iagnemma.
\newblock Self-supervised classification for planetary rover terrain sensing.
\newblock In \emph{Aerospace Conference, 2007 IEEE}, pages 1--9, March 2007.

\bibitem[Christe and Kottege(2016)]{christie2016icra}
J.~Christe and N.~Kottege.
\newblock Acoustics based terrain classification for legged robots.
\newblock In \emph{Proceedings of the IEEE International Conference on Robotics
  and Automation}, 2016.

\bibitem[Donahue et~al.(2015)Donahue, Hendricks, Guadarrama, Rohrbach,
  Venugopalan, Saenko, and Darrell]{donahue2014cvpr}
J.~Donahue, L.~A. Hendricks, S.~Guadarrama, M.~Rohrbach, S.~Venugopalan,
  K.~Saenko, and T.~Darrell.
\newblock Long-term recurrent convolutional networks for visual recognition and
  description.
\newblock In \emph{CVPR}, 2015.

\bibitem[Ellis(2007)]{ellis2007}
D.~P.~W. Ellis.
\newblock Classifying music audio with timbral and chroma features.
\newblock In \emph{8th International Conference on Music Information
  Retrieval}, 2007.

\bibitem[Eriksson et~al.(2008)Eriksson, Girod, Hull, Newton, Madden, and
  Balakrishnan]{eriksson2008mobsys}
J.~Eriksson, L.~Girod, B.~Hull, R.~Newton, S.~Madden, and H.~Balakrishnan.
\newblock The pothole patrol: Using a mobile sensor network for road surface
  monitoring.
\newblock In \emph{The Sixth Annual International conference on Mobile Systems,
  Applications and Services (MobiSys 2008)}, Breckenridge, U.S.A., June 2008.

\bibitem[Giannakopoulos et~al.(2006)Giannakopoulos, Kosmopoulos, Aristidou, and
  Theodoridis]{giannakopoulos2006setn}
T.~Giannakopoulos, D.~Kosmopoulos, A.~Aristidou, and S.~Theodoridis.
\newblock \emph{Advances in Artificial Intelligence: 4th Helenic Conference on
  AI, SETN 2006, Heraklion, Crete, Greece, May 18-20, 2006. Proceedings},
  chapter Violence Content Classification Using Audio Features, pages 502--507.
\newblock Springer Berlin Heidelberg, 2006.

\bibitem[Glorot and Bengio(2010)]{glorot2010aistats}
X.~Glorot and Y.~Bengio.
\newblock Understanding the difficulty of training deep feedforward neural
  networks.
\newblock In \emph{In Proceedings of the International Conference on Artificial
  Intelligence and Statistics (AISTATS’10). Society for Artificial
  Intelligence and Statistics}, 2010.

\bibitem[Graves(2013)]{graves13arxiv}
A.~Graves.
\newblock Generating sequences with recurrent neural networks.
\newblock \emph{arxiv preprint arxiv: 1308.0850}, 2013.

\bibitem[Graves and Jaitly(2014)]{graves2014icml}
A.~Graves and N.~Jaitly.
\newblock Towards end-to-end speech recognition with recurrent neural networks.
\newblock In T.~Jebara and E.~P. Xing, editors, \emph{Proceedings of the 31st
  International Conference on Machine Learning (ICML-14)}, pages 1764--1772,
  2014.

\bibitem[Hadsell et~al.(2009)Hadsell, Sermanet, Ben, Erkan, Scoffier,
  Kavukcuoglu, Muller, and LeCun]{hadsell2009jfr}
R.~Hadsell, P.~Sermanet, J.~Ben, A.~Erkan, M.~Scoffier, K.~Kavukcuoglu,
  U.~Muller, and Y.~LeCun.
\newblock Learning long-range vision for autonomous off-road driving.
\newblock \emph{Journal of Field Robotics}, 26\penalty0 (2):\penalty0 120--144,
  Feb. 2009.
\newblock ISSN 1556-4959.

\bibitem[Hinton et~al.(2012)Hinton, Srivastava, Krizhevsky, Sutskever, and
  Salakhutdinov]{hinton2012arxiv}
G.~E. Hinton, N.~Srivastava, A.~Krizhevsky, I.~Sutskever, and R.~Salakhutdinov.
\newblock Improving neural networks by preventing co-adaptation of feature
  detectors.
\newblock \emph{arXiv preprint arXiv:1207.0580}, 2012.

\bibitem[Hochreiter and Schmidhuber(1997)]{hochreiter1997lstm}
S.~Hochreiter and J.~Schmidhuber.
\newblock Long short-term memory.
\newblock \emph{Neural Computation}, 9\penalty0 (8):\penalty0 1735--1780, Nov.
  1997.

\bibitem[Hoepflinger et~al.(2010)Hoepflinger, Remy, Hutter, Spinello, and
  Siegwart]{hoepflinger2010}
M.~A. Hoepflinger, C.~D. Remy, M.~Hutter, L.~Spinello, and R.~Siegwart.
\newblock Haptic terrain classification for legged robots.
\newblock In \emph{2010 IEEE International Conference on Robotics and
  Automation}, pages 2828--2833, May 2010.

\bibitem[Jia et~al.(2014)Jia, Shelhamer, Donahue, Karayev, Long, Girshick,
  Guadarrama, and Darrell]{jia2014arxiv}
Y.~Jia, E.~Shelhamer, J.~Donahue, S.~Karayev, J.~Long, R.~Girshick,
  S.~Guadarrama, and T.~Darrell.
\newblock Caffe: Convolutional architecture for fast feature embedding.
\newblock \emph{arXiv preprint arXiv:1408.5093}, 2014.

\bibitem[Khunarsal et~al.(2013)Khunarsal, Lursinsap, and
  Raicharoen]{khunarsal2013jis}
P.~Khunarsal, C.~Lursinsap, and T.~Raicharoen.
\newblock Very short time environmental sound classification based on
  spectrogram pattern matching.
\newblock \emph{Information Sciences}, 243:\penalty0 57 -- 74, 2013.

\bibitem[Libby and Stentz(2012)]{libby2012icra}
J.~Libby and A.~T. Stentz.
\newblock Using sound to classify vehicle-terrain interactions in outdoor
  environments.
\newblock In \emph{Proceedings of the IEEE International Conference on Robotics
  and Automation}, May 2012.

\bibitem[Lin et~al.(2013)Lin, Chen, and Yan]{lin2013arxiv}
M.~Lin, Q.~Chen, and S.~Yan.
\newblock Network in network.
\newblock \emph{arXiv preprint arXiv:1312.4400}, 2013.

\bibitem[Loizou(2007)]{loizou2007speech}
P.~Loizou.
\newblock \emph{Speech Enhancement: Theory and Practice}.
\newblock Signal processing and communications. 2007.

\bibitem[Muller et~al.(2013)Muller, Jackel, LeCun, and Flepp]{muller2013spie}
U.~A. Muller, L.~D. Jackel, Y.~LeCun, and B.~Flepp.
\newblock Real-time adaptive off-road vehicle navigation and terrain
  classification.
\newblock \emph{Proc. SPIE}, 8741:\penalty0 87410A--87410A--19, 2013.

\bibitem[Namin et~al.(2014)Namin, Najafi, and Petersson]{namin2014iros}
S.~T. Namin, M.~Najafi, and L.~Petersson.
\newblock Multi-view terrain classification using panoramic imagery and lidar.
\newblock In \emph{Proceedings of the IEEE/RSJ International Conference on
  Intelligent Robots and Systems}, pages 4936--4943, Sept 2014.

\bibitem[Ojeda et~al.(2006)Ojeda, Borenstein, Witus, and Karlsen]{ojeda2006jfr}
L.~Ojeda, J.~Borenstein, G.~Witus, and R.~Karlsen.
\newblock Terrain characterization and classification with a mobile robot.
\newblock \emph{Journal of Field Robotics}, 23\penalty0 (2):\penalty0 103--122,
  2006.

\bibitem[Otte et~al.(2015)Otte, Laible, Hanten, Liwicki, and
  Zell]{otte2015esann}
S.~Otte, S.~Laible, R.~Hanten, M.~Liwicki, and A.~Zell.
\newblock Robust visual terrain classification with recurrent neural networks.
\newblock In \emph{In Proc.~of European Symposium on Artificial Neural
  Networks}, 2015.

\bibitem[Ozkul et~al.(2013)Ozkul, Saranli, and Yazicioglu]{cuneyitoglu2013mssp}
M.~C. Ozkul, A.~Saranli, and Y.~Yazicioglu.
\newblock Acoustic surface perception from naturally occurring step sounds of a
  dexterous hexapod robot.
\newblock \emph{Mechanical Systems and Signal Processing}, 40\penalty0
  (1):\penalty0 178--193, 2013.

\bibitem[Posner et~al.(2008)Posner, Cummins, and Newman]{posner2008rss}
I.~Posner, M.~Cummins, and P.~Newman.
\newblock Fast probabilistic labeling of city maps.
\newblock \emph{Proceedings of Robotics: Science and Systems IV, Zurich,
  Switzerland}, 2008.

\bibitem[Santamaria-Navarro et~al.(2015)Santamaria-Navarro, Teniente, Morta,
  and Andrade-Cetto]{angel2015jfr}
A.~Santamaria-Navarro, E.~H. Teniente, M.~Morta, and J.~Andrade-Cetto.
\newblock Terrain classification in complex three-dimensional outdoor
  environments.
\newblock \emph{Journal of Field Robotics}, 32\penalty0 (1):\penalty0 42--60,
  2015.

\bibitem[Snoek et~al.(2012)Snoek, Larochelle, and Adams]{snoek2012nips}
J.~Snoek, H.~Larochelle, and R.~P. Adams.
\newblock Practical bayesian optimization of machine learning algorithms.
\newblock In \emph{Advances in neural information processing systems}, pages
  2951--2959, 2012.

\bibitem[Suger et~al.(2015)Suger, Steder, and Burgard]{suger15icra}
B.~Suger, B.~Steder, and W.~Burgard.
\newblock Traversability analysis for mobile robots in outdoor environments: A
  semi-supervised learning approach based on 3d-lidar data.
\newblock In \emph{Proc.~of the IEEE Int.~Conf.~on Robotics \& Automation
  (ICRA)}, 2015.

\bibitem[Sung et~al.(2010)Sung, Kwak, and Lyou]{Sung2010jirs}
G.-Y. Sung, D.-M. Kwak, and J.~Lyou.
\newblock Neural network based terrain classification using wavelet features.
\newblock \emph{Journal of Intelligent \& Robotic Systems}, 59\penalty0
  (3):\penalty0 269--281, 2010.

\bibitem[Sutskever et~al.(2014)Sutskever, Vinyals, and Le]{sutskever2014nips}
I.~Sutskever, O.~Vinyals, and Q.~V. Le.
\newblock Sequence to sequence learning with neural networks.
\newblock In \emph{Advances in neural information processing systems}, pages
  3104--3112, 2014.

\bibitem[Thiemann et~al.(2013)Thiemann, Ito, and Vincent]{demand2013}
J.~Thiemann, N.~Ito, and E.~Vincent.
\newblock The diverse environments multi-channel acoustic noise database
  (demand): A database of multichannel environmental noise recordings.
\newblock In \emph{21st International Congress on Acoustics}, 2013.

\bibitem[Thrun et~al.(2006)Thrun, Montemerlo, Dahlkamp, Stavens, Aron, Diebel,
  Fong, Gale, Halpenny, Hoffmann, Lau, Oakley, Palatucci, Pratt, Stang,
  Strohband, Dupont, Jendrossek, Koelen, Markey, Rummel, van Niekerk, Jensen,
  Alessandrini, Bradski, Davies, Ettinger, Kaehler, Nefian, and
  Mahoney]{thrun2006jfr}
S.~Thrun, M.~Montemerlo, H.~Dahlkamp, D.~Stavens, A.~Aron, J.~Diebel, P.~Fong,
  J.~Gale, M.~Halpenny, G.~Hoffmann, K.~Lau, C.~Oakley, M.~Palatucci, V.~Pratt,
  P.~Stang, S.~Strohband, C.~Dupont, L.-E. Jendrossek, C.~Koelen, C.~Markey,
  C.~Rummel, J.~van Niekerk, E.~Jensen, P.~Alessandrini, G.~Bradski, B.~Davies,
  S.~Ettinger, A.~Kaehler, A.~Nefian, and P.~Mahoney.
\newblock Stanley: The robot that won the darpa grand challenge.
\newblock \emph{Journal of Field Robotics}, 23\penalty0 (9):\penalty0 661--692,
  2006.

\bibitem[Trautmann and Ray(2011)]{trautmann2011}
E.~Trautmann and L.~Ray.
\newblock Mobility characterization for autonomous mobile robots using machine
  learning.
\newblock \emph{Autonomous Robots}, 30\penalty0 (4):\penalty0 369--383, 2011.

\bibitem[Tzanetakis and Cook(2002)]{tzanetakis2002tsap}
G.~Tzanetakis and P.~Cook.
\newblock Musical genre classification of audio signals.
\newblock \emph{IEEE Transactions on Speech and Audio Processing}, 10\penalty0
  (5):\penalty0 293--302, Jul 2002.

\bibitem[Valada et~al.(2015)Valada, Spinello, and Burgard]{valada2015isrr}
A.~Valada, L.~Spinello, and W.~Burgard.
\newblock Deep feature learning for acoustics-based terrain classification.
\newblock In \emph{Proceedings of the International Symposium on Robotics
  Research}, 2015.

\bibitem[Verma(2008)]{verma2008pattern}
B.~Verma.
\newblock \emph{Pattern Recognition Technologies and Applications: Recent
  Advances: Recent Advances}.
\newblock IGI Global, 2008.

\bibitem[Weiss et~al.(2006)Weiss, Frohlich, and Zell]{weiss2006iros}
C.~Weiss, H.~Frohlich, and A.~Zell.
\newblock Vibration-based terrain classification using support vector machines.
\newblock In \emph{Proceedings of the IEEE/RSJ International Conference on
  Intelligent Robots and Systems}, pages 4429--4434, Oct 2006.

\bibitem[Wellman et~al.(1997)Wellman, Srour, and Hillis]{wellman1997spie}
M.~C. Wellman, N.~Srour, and D.~B. Hillis.
\newblock Feature extraction and fusion of acoustic and seismic sensors for
  target identification.
\newblock \emph{Proc. SPIE}, 3081:\penalty0 139--145, 1997.

\bibitem[Zaremba and Sutskever(2014)]{zaremba2014ariv}
W.~Zaremba and I.~Sutskever.
\newblock Learning to execute.
\newblock \emph{arxiv preprint arxiv: 1410.4615}, 2014.

\end{thebibliography}

\end{document}